\documentclass[runningheads]{llncs}

\usepackage{eccv}

\usepackage{eccvabbrv}

\usepackage{graphicx}
\usepackage{booktabs}
\usepackage{multirow}
\usepackage{color, colortbl}
\usepackage{tcolorbox}
\usepackage[accsupp]{axessibility}  

\definecolor{Gray}{gray}{0.9}

\newcommand{\como}{CoMo}

\usepackage[breaklinks,colorlinks]{hyperref}

\usepackage{orcidlink}

\begin{document}

\title{\como: Controllable Motion Generation \\through Language Guided Pose Code Editing} 


\author{Yiming Huang\inst{1}, Weilin Wan\inst{2}, Yue Yang\inst{1}, \\Chris Callison-Burch\inst{1}, Mark Yatskar\inst{1}, Lingjie Liu\inst{1}}

\authorrunning{Y. Huang et al.}
\titlerunning{Controllable Motion Generation}

\institute{University of Pennsylvania 
\and The University of Hong Kong\\
\email{ymhuang9@seas.upenn.edu}
}

\maketitle
\begin{abstract}
Text-to-motion models excel at efficient human motion generation, but existing approaches lack fine-grained controllability over the generation process. Consequently, modifying subtle postures within a motion or inserting new actions at specific moments remains a challenge, limiting the applicability of these methods in diverse scenarios. In light of these challenges, we introduce \textbf{CoMo}, a \textbf{Co}ntrollable \textbf{Mo}tion generation model, adept at accurately generating and editing motions by leveraging the knowledge priors of large language models (LLMs). Specifically, CoMo decomposes motions into discrete and semantically meaningful \textit{pose codes}, with each code encapsulating the semantics of a body part, representing elementary information such as ``left knee slightly bent''. Given textual inputs, CoMo autoregressively generates sequences of pose codes, which are then decoded into 3D motions. Leveraging pose codes as interpretable representations, an LLM can directly intervene in motion editing by adjusting the pose codes according to editing instructions. Experiments demonstrate that CoMo achieves competitive performance in motion generation compared to state-of-the-art models while, in human studies, CoMo substantially surpasses previous work in motion editing abilities. Project page: \url{https://yh2371.github.io/como/}.


\end{abstract}

\begin{figure}[t]
    \centering
    \includegraphics[width=0.9\textwidth]{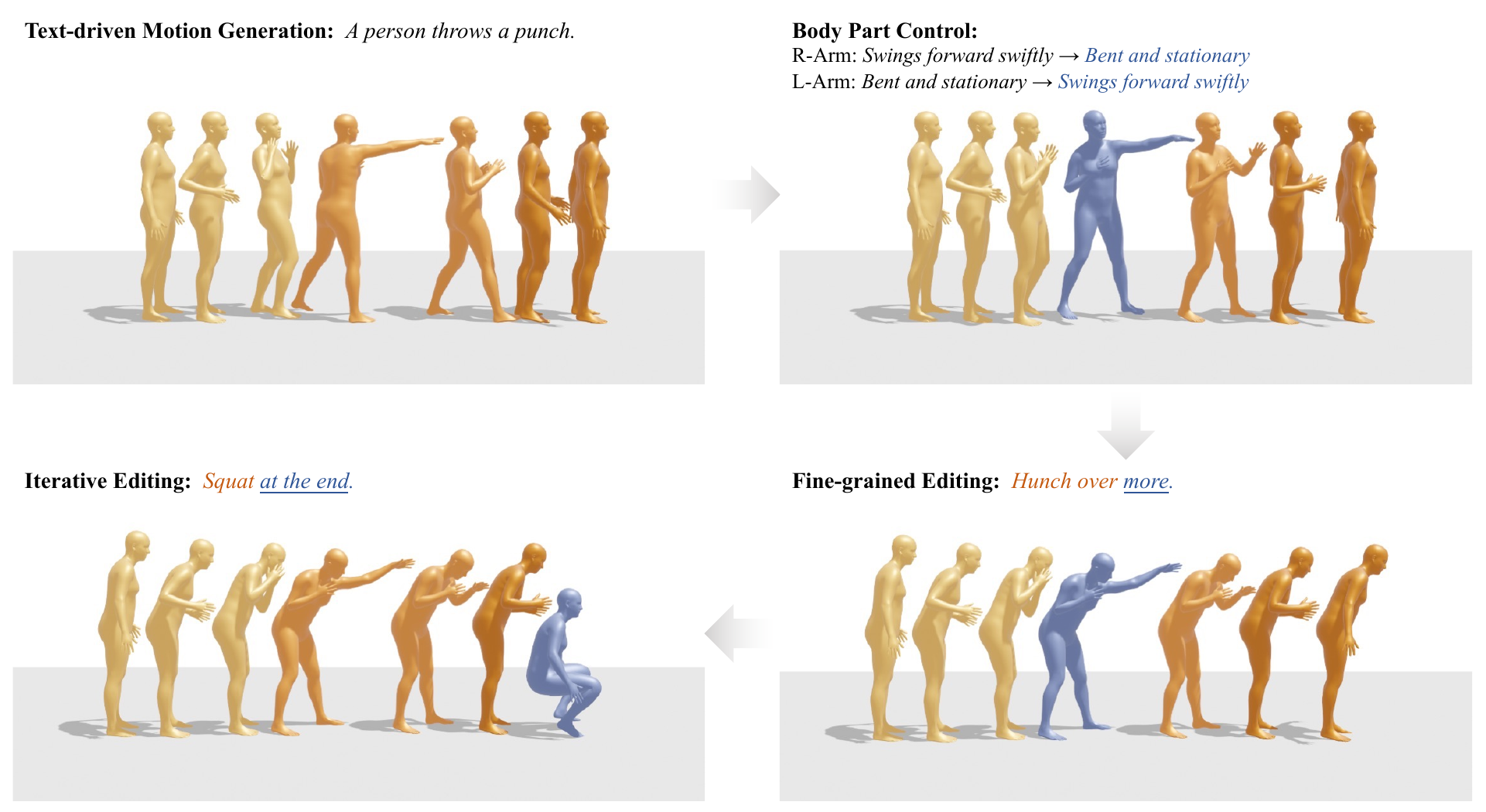} 
  \caption{CoMo, a language-guided human motion synthesis model, enables controllable generation from text inputs. CoMo allows for the control of individual body part movements, facilitates fine-grained editing of each joint and frame, and supports iterative editing that preserves the essence of the original motions.}
  \label{fig:teaser}
\end{figure}

\section{Introduction}
\label{sec:intro}
\noindent The diversity of natural and unconstrained human motion holds rich intricacies crucial for fostering a deeper understanding of human behavior. 
The synthesis of such motion is challenging because models must both create plausible dynamics and reason over many possible solutions to a specification. Various conditional signals have been explored for guiding human motion synthesis, including audio signals \cite{tseng2023edge,siyao2022bailando,zhu2023taming,yi2023generating} and simulated scenes \cite{couch,huang2023diffusion,Zhao:ICCV:2023}. Among these, natural language descriptions have emerged as a promising choice because they can communicate a broad set of needs naturally and could be used in design scenarios for animation and immersive technologies. 

Several pioneering works in the text-to-motion task have demonstrated the expressiveness of textual descriptions in effectively guiding the creation of human motion sequences \cite{t2m_guo,action2motion,guo2022tm2t,tevet2022motionclip, mdm, t2mgpt, jiang2024motiongpt, zhang2023motiongpt}. Approaches for this problem typically employ a method that maps a single text description to a latent code of attributes and then generates motion from codes. Such approaches may be poorly suited for fine-grained control of the generated motion.
The generations imprecisely correspond to the text because codes contain a superimposed representation of body parts that a generator must then disentangle. Furthermore,  capturing semantic relationships between text descriptions and low-level motion states may be too hard when the intermediate states must be simultaneously discovered.
The need to capture a broad range of textual inputs amplifies the problem.

To address these challenges, we propose the \textbf{Co}ntrollable \textbf{Mo}tion Generation model (\textbf{CoMo}). 
As shown in Figure \ref{fig:teaser}, CoMo can generate high-quality human motions from a broad range of text (e.g., \textit{``A person throws a punch.''}).
Users can control the generation by altering the descriptions for each body part, e.g., switching the left and right arm descriptions changes the punching hand. More importantly, CoMo enables detailed editing across frames and joints (e.g., \textit{``Hunch over more.''}), adding actions (e.g., \textit{``Squat at the end.''}), and varying speed and even emotion (e.g., \textit{``more dramatic''}), as depicted in Figure \ref{fig:qualitative}.

CoMo achieves these capabilities by representing motions as interpretable ``pose codes'', with each code defining the state of a specific body part at a given moment, e.g., \textit{``right arm straight''}. As demonstrated in Figure \ref{fig:motion_generation}, our method starts by factorizing a motion sequence into a series of temporal states, each composed of pose codes that describe the motion's kinematic characteristics. This structure allows for precise encoding of human motion sequences in time and fine-grained control of kinematic joint states.
Since these encodings are explicitly interpretable, people can interact with and modify the sequences intuitively.
We show that this process can also be made instructable with natural language by allowing large language models (LLM) to edit the codes as well.

Leveraging pose codes as interpretable motion representations, the three main components of CoMo work jointly to effectively generate and edit motion: (1) The \textit{Motion Encoder-Decoder} applies heuristic rules to parse motions into sequences of semantically meaningful pose codes and trains a decoder to reconstruct these codes back into motions; (2) The \textit{Motion Generator}, a transformer-based model, generates pose codes conditioned on text inputs and LLM-generated fine-grained descriptions; (3) The \textit{Motion Editor} uses LLMs to modify and refine pose code sequences based on editing instructions. The resulting pose code sequences, whether generated or edited, are subsequently decoded into motion sequences using the previously trained decoder. CoMo allows for intuitive, language-controlled adjustments to the motion sequences, both temporally and kinematically, closely aligning the generated motions with users' creative intentions and the nuances expressed in their textual descriptions, making the process user-friendly and adaptable to diverse applications.

We evaluate the effectiveness of CoMo in text-driven motion generation against state-of-the-art methods on the HumanML3D~\cite{t2m_guo} and KIT~\cite{kitdataset} datasets, ranking within the top 3 across most metrics. Beyond the competitive motion generation capabilities of CoMo, we also conducted a human evaluation with 54 participants for motion editing. On average, over 70\% of annotators preferred the editing results produced by CoMo. CoMo's motion editing abilities allow for potential new applications, such as dialog-based motion generation.

In summary, our contributions in this paper are threefold:
\begin{enumerate}
    \item We propose a semantic motion representation that factorizes motion sequences across space and time into explicit and interpretable pose codes.
    \item We present a transformer-based model that autoregressively generates sets of low-level pose codes conditioned upon the high-level text description and fine-grained, body-part-specific descriptions generated by LLMs.
    \item We demonstrate the capability of using the semantic low-level pose codes as an intuitive motion editing interface for LLMs.
\end{enumerate} 

\section{Related Work}
\textbf{Text Conditioned Human Motion Generation.} Conditional motion synthesis involves an interactive process for generating diverse, human-like motion from multi-modal user input, including text descriptions \cite{t2m_guo, mdm, priormdm, MLD, t2mgpt, jiang2024motiongpt, zhang2023motiongpt, motionDiffuse, flame, temos, tevet2022motionclip, Zhou_2023_CVPR, zhou2023emdm}, action categories \cite{action2motion, actiongpt, TEACH} and physics-based signals \cite{yuan2022physdiff, diffmimic, dou2023c}. To tackle the challenges in effectively mapping the intricacies of textual descriptions to meaningful movements, building a shared latent space has been a widely adopted solution \cite{temos, TEACH, tevet2022motionclip,tevet2022motionclip, wan2023diffusionphase}. Inspired by successful applications in image generation, the Vector Quantized Variational Autoencoder (VQ-VAE) model \cite{vqvae} has been widely applied to represent motion as discretized tokens, which can then be effectively combined with autoregressive transformer architectures for producing coherent motion sequences \cite{guo2022tm2t,t2mgpt, Zhou_2023_CVPR, jiang2024motiongpt, zhang2023motiongpt}. Conditional diffusion models are also increasingly powerful for generating high-fidelity results due to their capability of modeling complex distributions \cite{yuan2022physdiff, motionDiffuse, flame, remodiffuse, mdm, MLD, priormdm, 2023mofusion, xie2023omnicontrol}.\\

\noindent \textbf{Fine-grained motion generation} has gained significant interest due to its extensive practical applications. However, current methods have not fully integrated spatial and temporal details.
TEACH \cite{TEACH} demonstrates improvements in motion smoothness by incorporating extensive temporal annotations but requires additional labeled data and overlooks lower-level spatial details.
\cite{actiongpt, shi2023generating, SINC:ICCV:2022} leverage LLMs to generate detailed text descriptions for whole-body motions and individual body parts, aiming to map text to spatiotemporal details. However, without explicit supervision, these fine-grained details may not align with the respective motion. GraphMotion \cite{jin2023act} proposes a hierarchical semantic graph that enforces coarse-to-fine topology within text-to-motion diffusion. 
However, constructing the hierarchical graph depends on the semantic parsing of text details and thus may struggle for ambiguous inputs. Also, the graph nodes do not model the lower-level spatial characteristics and relations.
To this end, we propose body-part-specific semantic pose codes for achieving fine-grained representation of motion sequences.
In addition, these pose codes are used as context to guide LLMs in generating motion-coherent descriptions for different body parts to enhance the connection between text and fine-grained motion details.\\

\noindent \textbf{Motion Editing} enables users to interactively refine generated motions to suit their expectations. PoseFix \cite{delmas2023posefix} automates 3D pose and text modifier generation for supervised editing, but frame-wise pose edits lack efficiency and temporal consistency.
TLControl \cite{wan2023tlcontrol} allows motion editing using joint-level trajectories but relies on high-quality trajectory inputs and is not intuitive for interactive motion editing.
\cite{deepsynthesis} employs an autoencoder to optimize a motion manifold under positional, bone length, and trajectory constraints for smooth edits.
Diffusion-based approaches \cite{flame,goel2023iterative,xie2023omnicontrol} can achieve zero-shot spatiotemporal editing by infilling particular joints or frames, which may form unnatural discontinuities.
Recently, FineMoGen \cite{zhang2023finemogen} optimizes global attention for fine-grained editing but generates new sequences for each edit, limiting motion consistency. In our approach, we encode motions into semantic pose codes, which serve as context to prompt an LLM to edit the original motion directly, encouraging motion consistency.

\section{Method}
\begin{figure*}
    \centering
    \includegraphics[width=\textwidth]{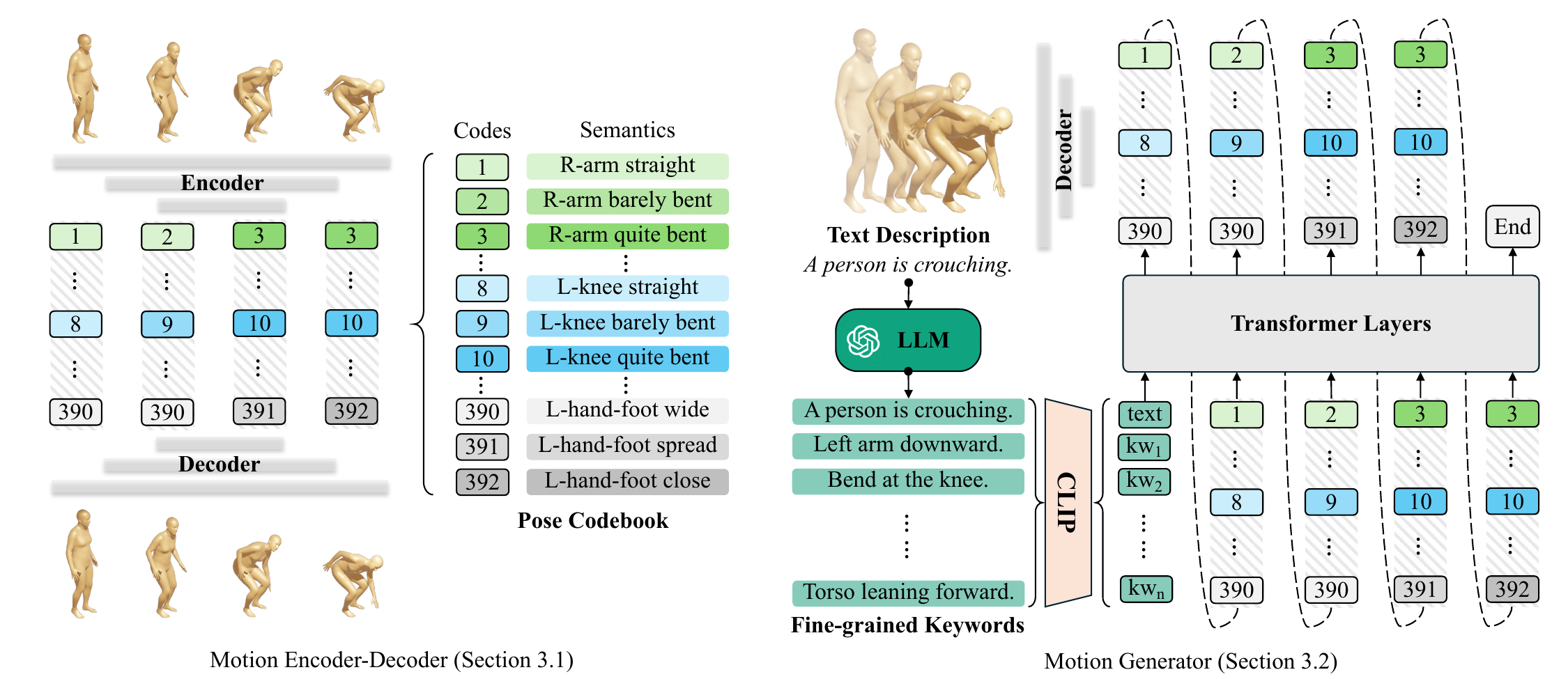}
    \caption{\textbf{Overview of CoMo for text-driven motion generation.} \textbf{Motion Encoder-Decoder} (left) utilizes a predefined codebook to encode motions into pose codes and learns a decoder to reconstruct the motions. \textbf{Motion Generator} (right), a transformer-based model, predicts pose codes autoregressively, conditioned on the text descriptions and LLM-generated fine-grained keywords. The generated pose codes are then decoded back into motions using the previously trained decoder.}
    \label{fig:motion_generation}
\end{figure*}

CoMo is a unified framework for fine-grained, text-driven human motion generation and editing. Figures \ref{fig:motion_generation} and \ref{fig:motion_editing} present an overview of CoMo, which consists of three key components: 1) \textbf{Motion Encoder-Decoder} (Sec. \ref{sec:codebook}) decomposes motions into sequences of pose codes. These codes are then mapped back to motions through a decoder; 2) \textbf{Motion Generator} (Sec. \ref{sec:generator}) generates sequences of pose codes given high-level text descriptions and LLM-generated fine-grained keywords; 3) \textbf{Motion Editor} (Sec. \ref{sec:edit}) employs an LLM to perform modifications on pose codes in a zero-shot manner.

\subsection{Motion Encoder-Decoder}
\label{sec:codebook}

As shown in the left panel of Figure \ref{fig:motion_generation}, given $T$ frames of motion $X=\left\{x_i\right\}_{i=1}^T$, the motion encoder $\mathcal{E}$ projects them into a sequence of pose codes using a codebook, denoted as $Z = \mathcal{E}\left(X\right)$, where $Z = \left\{z_i\right\}_{i=1}^L$. Here, $L = T / l$ represents the length of the pose code sequence, and $l$ is the downsampling rate over poses. The decoder $\mathcal{D}$ then decodes the pose codes back into motions, expressed as $X_{\text{rec}} = \mathcal{D}(\hat{Z})$, where $\hat{Z}$ stands for the latent features of pose codes. In the following, we will explain how to build the pose codebook, encode the motions using these codes, and learn the decoder to recover the motions.\\

\noindent \textbf{Pose Codebook.} Unlike \cite{t2mgpt}, which uses an autoencoder to obtain implicit motion representations, CoMo predefines a semantic pose codebook to achieve interpretable motion factorization across time and space. Following PoseScript \cite{delmas2022posescript}, we construct a pose codebook with $N$ codes, $\mathcal{C} = \left\{c_n\right\}_{n=1}^N$, with $c_n \in \mathbb{R}^{d_c}$, where $d_c$ is the dimension of the learnable codebook entry. Each code $c_n$ is associated with a semantic meaning, representing a state of a body part or spatial relationships between body parts, e.g., ``left knee slightly bent'', ``left hand and left foot close''. These codes are further grouped into $K$ pose categories, each encompassing different states of the same body parts, e.g., the pose category ``left knee angle'' includes codes like ``left knee slightly bent'', ``left knee partially bent'',  etc.\\

\noindent \textbf{Motion Encoder.} Given the codebook $\mathcal{C}$ with $N$ pose codes grouped into $K$ categories, we encode motions into $K$-hot $N$ dimensional vectors, denoted as $Z = \mathcal{E}\left(X\right) = \left\{z_{i}\right\}_{i=1}^{L}$, where $z_{i} \in \mathbb{R}^{N}$. Here, K-hot indicates that $K$ elements of $z_i$ are set to $1$, while the others are $0$. Specifically, at each time step $i$, we enforce mutual exclusivity between codes within the same category so that only one pose code per pose category is activated (set to $1$ in the vector). To determine whether a pose code $c \in \mathcal{C}$ applies to a motion $x$, we use an off-the-shelf skeleton parser \cite{delmas2022posescript}, denoted as $\mathcal{P}(c, x) \rightarrow \{0, 1\}$. The parser $\mathcal{P}$ analyzes the 3D joint positions of a skeleton in SMPL format \cite{SMPL:2015}, evaluating whether the pose meets specific heuristic threshold conditions for a given pose code. For instance, if the angle formed by the left shoulder, elbow, and wrist joints is less than 20 degrees, the code ``left arm completely bent'' is true. Therefore, instead of training an encoder model, we can explicitly factorize the motion sequence as:
\begin{equation}
    Z = \mathcal{E}\left(X\right) = \left\{\left\{\mathcal{P}\left(c_{n}, x_{i \times l}\right)\right\}_{n=1}^{N}\right\}_{i=1}^{L}
\end{equation}
where $x_{i \times l}$ is the motion frame extracted at the sampling rate $l$.\\

\noindent \textbf{Motion Decoder.} To develop a meaningful codebook, we train a 1D convolutional decoder \cite{t2mgpt}, denoted as $\mathcal{D}$, over the latent features $\hat{Z}$ to reconstruct the original motion sequence, expressed as $X_{\text{rec}} = \mathcal{D}(\hat{Z})$. The latent features $\hat{Z} \in \mathbb{R}^{L \times d_c}$ are derived by summing the active codebook entries $c_n \in \mathcal{C}$, as indicated by the $K$-hot vector:
\begin{equation}
    \hat{Z} = \left\{\sum_{n=1}^{N}\mathcal{P}\left(c_n, x_{i \times l}\right) \cdot c_{n}\right\}_{i=1}^{L}
\end{equation}

\noindent We define $V(X) = \left\{x_{i+1}-x_{i}\right\}_{i=1}^{T-1}$ as the velocity of the $T$-frame motion sequence $X$. The reconstruction objective is formulated with smooth L1 loss $\mathcal{L}_1$:
\begin{equation}
    \mathcal{L}_{\text{rec}} = \mathcal{L}_{1}\left(X,X_\text{rec}\right)+ \lambda \cdot \mathcal{L}_{1}\left(V\left(X\right), V\left(X_\text{rec}\right)\right)
\end{equation}
where the hyperparameter $\lambda$\footnote{Following \cite{t2mgpt}, we set $\lambda$ to 0.5.} balances the velocity and motion loss. The learned codebook and decoder are then frozen for motion generation and editing.

\subsection{Motion Generator} \label{sec:generator}
As illustrated in the right panel of Figure \ref{fig:motion_generation}, the Motion Generator, conditioned on text input, aims to generate a sequence of pose codes, which will be decoded into motions. Utilizing the learned pose codebook, we map a motion sequence $X = \{x_i\}_{i=1}^T$ to a sequence of $K$-hot, $N$-dimensional vectors $Z^{1:L} = \left\{\{z_i^n\}_{n=1}^{N}\right\}_{i=1}^{L}$, where $z_i^n$ is an indicator function that is activated if the corresponding pose code $c_n$ is true at the time index $i$. To denote the end of a motion sequence, we append an \texttt{<End>} code to each latent vector, activated when the motion stops, and the dimension of each latent vector $z_i$ becomes $N+1$.

Treating the true label of each indicator $z_i^n$ as an independent Bernoulli random variable, the text-to-motion generation task can be framed as an autoregressive multi-label prediction problem. Given the previous $K$-hot vectors $Z^{1:i-1}$ and a text condition $t$, our goal is to predict the Bernoulli distributions for the indicator terms $z_i$ at the next time step $i$, represented as $P\left(z_i | t, Z^{1:i-1}\right)$.

To achieve this goal, we adopt a decoder-only transformer architecture with causal self-attention \cite{t2mgpt}.\footnote{Following the method for processing image patches in Vision Transformers \cite{dosovitskiy2021an}, a linear layer projects the $K$-hot vectors before input into the transformer architecture.} The likelihood of the full sequence is: 
\begin{equation}
    P(Z|t) = \prod_{i=1}^{L} \prod_{n=1}^{N+1}p\left(z_i^n\,|\,t,z_{1:i-1}^{1:N+1}\right)
\end{equation}
We implement a binary cross-entropy loss and aim to maximize the average log-likelihood across all Bernoulli distributions:
\begin{equation}
    \mathcal{L}_\text{gen} = -\frac{1}{L(N+1)}\sum_{i=1}^{L}\sum_{n=1}^{N+1}\mathbb{E}_{{z_i^n} \sim Ber\left(z_i^n\right)}\left[\log p\left(z_i^n\,|\,t, z_{1:i-1}^{1:N+1}\right)\right]
\end{equation}

\noindent The predicted sequences of $K$-hot vectors are mapped back to respective pose codes, which can then be decoded to motion sequences through the decoder $\mathcal{D}$.\\

\noindent \textbf{Fine-grained Keywords.} To help the model capture more fine-grained details, we enhance the text description by using GPT-4 \cite{openai2023gpt4} to generate a keyword for each of the 10 body parts\footnote{The 10 body parts are \textit{head}, \textit{torso}, \textit{left arm}, \textit{right arm}, \textit{left hand}, \textit{right hand}, \textit{left leg}, \textit{right leg}, \textit{left feet}, \textit{right feet}.} and a keyword to describe the overall \textit{mood} of the motion.
For example, as shown in Figure \ref{fig:motion_generation}, GPT-4 generates ``bend at the knee'', ``torso leaning forward'', etc., as the keywords for the motion \textit{``crouching''}. We use CLIP \cite{radford2021learning} to extract text embeddings of the keywords and original description. The embedding of the original text, followed by the embeddings of 11 keywords, form the initial token sequence that conditions the subsequent motion generation. The effectiveness of using these keywords to improve generation performance is validated in Table \ref{tab:finegrained}.

\subsection{Motion Editor} \label{sec:edit}
\begin{figure*}[!t]
    \centering
    \includegraphics[width=\textwidth]{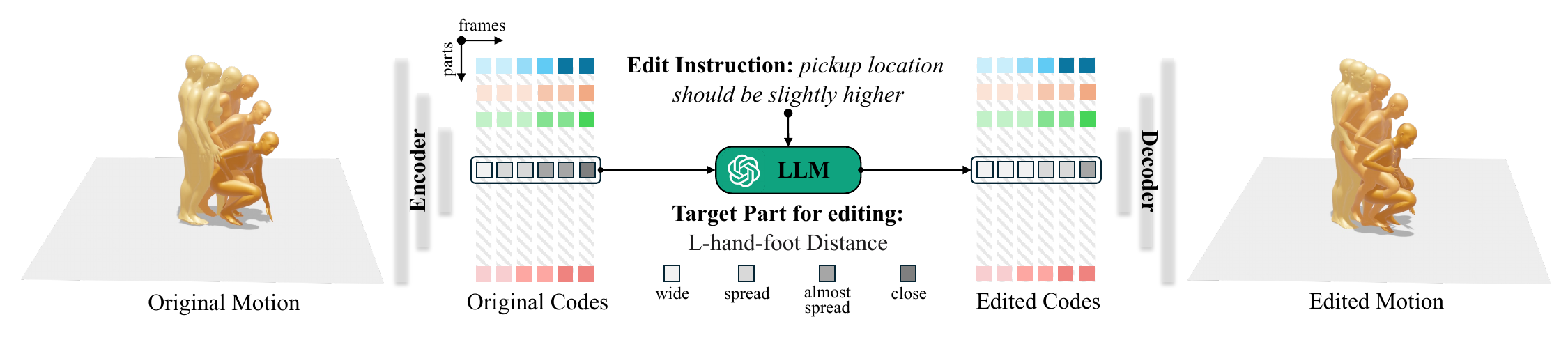}
    \caption{\textbf{Overview of CoMo for Fine-Grained Motion Editing:} Given an original motion and an editing instruction, CoMo encodes the motion into pose codes, serving as the context to prompt an LLM. The LLM identifies the target codes for editing based on the instructions and updates the corresponding codes accordingly. These edited codes are then decoded back into motions to satisfy the user's requirements.}
    \label{fig:motion_editing}
\end{figure*}

As illustrated in Figure \ref{fig:motion_editing}, given an original motion, such as \textit{crouching}, and an editing instruction like ``pickup location should be slightly higher'', the Motion Editor modifies the original motion to satisfy the requirements. Benefiting from our approach of encoding motions into explicit semantic pose codes, a Large Language Model (LLM) can interpret the motion and utilize its knowledge to reason about and execute editing instructions on an encoded motion sequence. 

To simplify the task for the LLM, we design a sequential prompting\footnote{The complete prompts we use are available in the Appendix.} strategy that consists of three steps:
\smallbreak
\noindent (1) \textbf{Identify the frames for editing.} The LLM or user determines the start and end indices of the motion segment where editing is needed. The identified subset of frames is retrieved and passed to the next step.
\smallbreak
\noindent (2) \textbf{Identify the body parts for editing.} The LLM identifies which body parts require editing. The subset of corresponding pose categories is then retrieved. For example, in Figure \ref{fig:motion_editing}, the category ``L-hand-foot distance'' is selected for modification to satisfy the editing instruction of ``pickup location higher''.
\smallbreak
\noindent (3) \textbf{Edit the pose codes.} The LLM reviews each selected pose category to decide how the pose codes should be altered to align with the editing instructions. For instance, as depicted in Figure \ref{fig:motion_editing}, the pose codes change from \textit{close} to \textit{almost spread} to mirror the instruction ``pickup location should be slightly higher''.
\smallbreak
The edited segments of the pose codes are seamlessly integrated with their unedited counterparts, forming the complete edited sequence of pose codes, which is passed through the decoder $\mathcal{D}$ to reconstruct the final edited motion.

\section{Evaluation}

In this section, we evaluate CoMo from two perspectives: (1) experiments to demonstrate that CoMo achieves state-of-the-art performance on text-driven \textbf{motion generation} (Sec \ref{sec: motion_generation}), and (2) human evaluation to showcase that CoMo is superior over existing methods in \textbf{motion editing} (Sec \ref{sec: human evaluation}). We also conduct comprehensive ablation studies (Sec \ref{sec: ablation}) to validate our model design.

\subsection{Experiments on Motion Generation} \label{sec: motion_generation}
\noindent \textbf{Datasets.} We evaluate our approach on two standard datasets for text-driven motion generation: HumanML3D~\cite{t2m_guo} and KIT Motion Language (KIT-ML)~\cite{kitdataset}. HumanML3D is a large-scale, diverse collection of human motion, including 14,616 distinct human motion capture sequences alongside 44,970 textual descriptions composed of 5,371 distinct words. The motion sequences are extracted from the HumanAct12 \cite{action2motion} and AMASS \cite{AMASS:ICCV:2019} datasets and preprocessed to 20 FPS. KIT-ML consists of 3,911 human motion sequences annotated with 6,278 distinct text annotations, forming a total vocabulary size of 1,623. The motion sequences are extracted from the KIT \cite{Mandery2016b} and CMU \cite{cmu_mocap} motion databases with a frame rate of 12.5 FPS. The motion sequences in KIT-ML and HumanML3D are all padded to 196 frames in length for training. Both datasets are split into $80\%$ training, $5\%$ validation, and $15\%$ test sets as designated in \cite{t2m_guo}.\\

\noindent \textbf{Evaluation Metrics.} We adhere to the protocol proposed in \cite{t2m_guo} and select the following five performance metrics for evaluation: \textit{Frechet Inception Distance (FID)} measures the similarity between generated and real motion sequences. \textit{R-Precision} and \textit{Multimodal Distances (MM-DIST)} assess the relevance of generated motion sequences to their corresponding text descriptions. \textit{Diversity} and \textit{Multimodality (MModality)} reflect the variability of the generated motion sequences. We employ the pre-trained network from \cite{t2m_guo} to derive the motion and text feature embeddings necessary for calculating these metrics.\footnote{Details of metric calculations are provided in the Appendix.}\\

\noindent \textbf{Hyperparameters.} In line with \cite{t2m_guo}, the motion sequences from KIT-ML and HumanML3D are transformed into motion features with dimensions of $261$ and $263$, using $21$ and $22$ SMPL joints, respectively. These features represent global translations and rotations and local joint positions, velocities, and rotations. Employing PoseScript \cite{delmas2022posescript} as the skeleton parser, we define $70$ pose categories encompassing $392$ pose codes. The codebook size is $392 \times 512$.\footnote{The definitions of all pose codes are listed in the Appendix.} The downsampling rate $l$ is set to $4$, and the maximum length of the code sequence produced by the Motion Generator is $50$. All hyperparameters are tuned using the HumanML3D validation set (see ablation studies in Sec \ref{sec: ablation}).\\
\begin{table}[!t]
\centering
\caption{Comparison with the state-of-the-art methods on HumanML3D~\cite{t2m_guo} test set. The best performance is \textbf{bold}, and the second best is \underline{underlined}.}
\resizebox{\textwidth}{!}{%
\begin{tabular}{lcccccccc}
\toprule
\multirow{2}{*}{\textbf{Method}}     & \multicolumn{3}{c}{\textbf{R-Precision $\uparrow$}} & \multirow{2}{*}{\textbf{FID $\downarrow$}} & \multirow{2}{*}{\textbf{MM-DIST $\downarrow$}} & \multirow{2}{*}{\textbf{Diversity $\uparrow$}} & \multirow{2}{*}{\textbf{MModality $\uparrow$}}\\ \cmidrule{2-4} & \textbf{Top-1} & \textbf{Top-2} & \textbf{Top-3} & & & & \\ \midrule
Real Motion & $0.511^{\pm .003}$ & $0.703^{\pm.003}$ & $0.797^{\pm.002}$ & $0.002^{\pm.000}$ & $2.974^{\pm .008}$ & $9.503^{\pm .085}$  & -\\
CoMo Recons. & $0.508^{\pm.002}$ & $0.697^{\pm.002}$ & $0.792^{\pm.002}$ & $0.041^{\pm.000}$ & $3.003^{\pm.006}$ & $9.563^{\pm .100}$ & -\\ \midrule
Guo et al. \cite{t2m_guo} & $0.457^{\pm .002}$ & $0.639^{\pm .003}$ & $0.740^{\pm.003}$ & $1.067^{\pm.002}$ & $3.340^{\pm.008}$ & $9.188^{\pm.002}$ & $2.090^{\pm.083}$\\
TM2T \cite{guo2022tm2t} & $0.424^{\pm .002}$ & $0.618^{\pm .003}$ & $0.729^{\pm.002}$ & $1.501^{\pm.017}$ & $3.467^{\pm.011}$ & $8.589^{\pm .086}$ & $2.424^{\pm.093}$ \\
TEMOS \cite{temos} & $0.424^{\pm .002}$ & $0.612^{\pm .002}$ & $0.722^{\pm.002}$ & $3.734^{\pm.028}$ & $3.703^{\pm.008}$ & $8.973^{\pm.071}$ & $0.368^{\pm.018}$ \\
MDM \cite{mdm} & $0.320^{\pm .005}$ & $0.498^{\pm .004}$ & $0.611^{\pm.007}$ & $0.544^{\pm.044}$ & $5.566^{\pm.027}$ & $9.559^{\pm.086}$ & $\textbf{2.799}^{\pm.072}$\\
MotionDiffuse \cite{motionDiffuse} & $0.491^{\pm .001}$ & $0.681^{\pm .001}$ & $0.782^{\pm.001}$ & $0.630^{\pm.001}$ & $3.113^{\pm.001}$ & $9.410^{\pm.049}$ & $1.533^{\pm.042}$\\
MLD \cite{MLD} & $0.481^{\pm .003}$ & $0.673^{\pm .003}$ & $0.772^{\pm.002}$ & $0.473^{\pm.013}$ & $3.196^{\pm.010}$ & $9.724^{\pm.082}$ & $2.413^{\pm.079}$\\
T2M-GPT \cite{t2mgpt} & $0.491^{\pm .001}$ & $0.680^{\pm .003}$ & $0.775^{\pm.002}$ & $\textbf{0.116}^{\pm.004}$ & $3.118^{\pm.011}$ & $\underline{9.761}^{\pm.081}$ & $1.831^{\pm.048}$\\
MotionGPT \cite{jiang2024motiongpt} & $0.492^{\pm .003}$ & $0.681^{\pm .003}$ & $0.778^{\pm.002}$ & $0.232^{\pm.008}$ & $3.096^{\pm.008}$ & $9.528^{\pm.071}$ & $2.008^{\pm.084}$\\
GraphMotion \cite{jin2023act} & $\textbf{0.504}^{\pm .003}$ & $\textbf{0.699}^{\pm .002}$ & $\underline{0.785}^{\pm.002}$ & $\textbf{0.116}^{\pm.007}$ & $3.070^{\pm.008}$ & $9.692^{\pm.067}$ & $\underline{2.766}^{\pm.096}$ \\
FineMoGen \cite{zhang2023finemogen} & $\textbf{0.504}^{\pm .003}$ & $0.690^{\pm .002}$ & $0.784^{\pm.002}$ & $\underline{0.151}^{\pm.008}$ & $\textbf{2.998}^{\pm.008}$ & $9.263^{\pm.067}$ & $2.696^{\pm.079}$ \\ \midrule
CoMo (Ours)       & $\underline{0.502}^{\pm .002}$ & $\underline{0.692}^{\pm .007}$ & $\textbf{0.790}^{\pm .002}$ & $0.262^{\pm .004}$ & $\underline{3.032}^{\pm .015}$ & $\textbf{9.936}^{\pm .066}$ & $1.013^{\pm.046}$\\
\bottomrule
\end{tabular}
}
\label{tab:humanml results}
\end{table}


\begin{table}[!t]
\centering
\caption{Comparison with the state-of-the-art methods on KIT~\cite{kitdataset} test set. The best performance is \textbf{bold}, and the second best is \underline{underlined}, the third best is \textit{italic}.}
\resizebox{\textwidth}{!}{%
\begin{tabular}{lccccccccc}
\toprule
\multirow{2}{*}{\textbf{Method}}     & \multicolumn{3}{c}{\textbf{R-Precision $\uparrow$}} & \multirow{2}{*}{\textbf{FID $\downarrow$}} & \multirow{2}{*}{\textbf{MM-DIST $\downarrow$}} & \multirow{2}{*}{\textbf{Diversity $\uparrow$}} & \multirow{2}{*}{\textbf{MModality $\uparrow$}} \\ \cmidrule{2-4} & \textbf{Top-1} & \textbf{Top-2} & \textbf{Top-3} & & & & \\ \midrule
Real Motion & $0.424^{\pm .005}$ & $0.649^{\pm.006}$ & $0.779^{\pm.006}$ & $0.031^{\pm.006}$ & $2.788^{\pm .012}$ & $11.08^{\pm .097}$ & -\\
CoMo Recons. & $0.387^{\pm.005}$ & $0.603^{\pm.005}$ & $0.730^{\pm.005}$ & $0.254^{\pm.007}$ & $3.046^{\pm.011}$ & $10.73^{\pm .128}$ & -\\ \midrule
Guo et al. \cite{t2m_guo} & $0.370^{\pm .005}$ & $0.569^{\pm .007}$ & $0.693^{\pm.007}$ & $2.770^{\pm.109}$ & $3.401^{\pm.008}$ & $10.91^{\pm.119}$ & $1.482^{\pm.065}$\\
TM2T \cite{guo2022tm2t} & $0.280^{\pm .005}$ & $0.463^{\pm .006}$ & $0.587^{\pm.005}$ & $3.599^{\pm.153}$ & $4.591^{\pm.026}$ & $9.473^{\pm .117}$ & $\underline{3.292}^{\pm.081}$\\
TEMOS \cite{temos} & $0.370^{\pm .005}$ & $0.569^{\pm .007}$ & $0.693^{\pm.007}$ & $2.770^{\pm.109}$ & $3.401^{\pm.008}$ & $10.91^{\pm.119}$ & $0.532^{\pm.034}$ \\
MDM \cite{mdm} & $0.164^{\pm .004}$ & $0.291^{\pm .004}$ & $0.396^{\pm.004}$ & $0.497^{\pm.021}$ & $9.191^{\pm.022}$ & $10.85^{\pm.109}$ & $1.907^{\pm.214}$\\
MotionDiffuse \cite{motionDiffuse} & $0.417^{\pm .004}$ & $0.621^{\pm .004}$ & $0.739^{\pm.004}$ & $1.954^{\pm.062}$ & $\textit{2.958}^{\pm.005}$ & $\underline{11.10}^{\pm.143}$ & $0.730^{\pm.013}$\\
MLD \cite{MLD} & $0.390^{\pm .008}$ & $0.609^{\pm .008}$ & $0.734^{\pm.007}$ & $0.404^{\pm.027}$ & $3.204^{\pm.027}$ & $10.80^{\pm.117}$ & $2.192^{\pm.071}$\\
T2M-GPT \cite{t2mgpt} & $0.416^{\pm .006}$ & $0.627^{\pm .006}$ & $\textit{0.745}^{\pm.006}$ & $0.514^{\pm.029}$ & $3.007^{\pm.023}$ & $10.92^{\pm.108}$ & $1.570^{\pm.039}$\\
MotionGPT \cite{jiang2024motiongpt} & $0.366^{\pm .005}$ & $0.558^{\pm .004}$ & $0.680^{\pm.005}$ & $0.510^{\pm.016}$ & $3.527^{\pm.021}$ & $10.35^{\pm.084}$  & $\textit{2.328}^{\pm.117}$\\
GraphMotion \cite{jin2023act} & $\textit{0.429}^{\pm .007}$ & $\underline{0.648}^{\pm .006}$ & $\underline{0.769}^{\pm.008}$ & $\underline
{0.313}^{\pm.013}$ & $3.076^{\pm.022}$ & $\textbf{11.12}^{\pm.135}$ & $\textbf{3.627}^{\pm.113}$\\
FineMoGen \cite{zhang2023finemogen} & $\textbf{0.432}^{\pm .006}$ & $\textbf{0.649}^{\pm .005}$ & $\textbf{0.772}^{\pm.008}$ & $\textbf{0.178}^{\pm.007}$ & $\textbf{2.869}^{\pm.014}$ & $10.85^{\pm.115}$ & $1.877^{\pm.093}$\\ \midrule
CoMo (Ours)       & $\underline{0.422}^{\pm .009}$ & $\textit{0.638}^{\pm .007}$ & $\textit{0.765}^{\pm .011}$ & $\textit{0.332}^{\pm .045}$ & $\underline{2.873}^{\pm .021}$ & $\textit{10.95}^{\pm .196}$ & $1.249^{\pm.008}$\\
\bottomrule
\end{tabular}
}
\label{tab:kit results}
\end{table}

\noindent \textbf{Training Details.} We apply GPT-4 (\texttt{gpt-4-0613}) \cite{openai2023gpt4} to generate fine-grained descriptions and a frozen CLIP ViT-B/32 \cite{radford2021learning} to encode text. We employ the AdamW optimizer for training. The Motion Decoder is trained for 200K iterations with a $10^{-4}$ learning rate and batch size 256. The Motion Generator is trained for 300K iterations with a learning rate of $10^{-4}$ and batch size of 64. The Motion Decoder and Motion Generator are trained on a single NVIDIA RTX A6000 GPU for approximately 9 and 60 hours, respectively. The model with the best FID score on the validation split of each dataset is retained for testing.\\

\noindent \textbf{CoMo achieves competitive results on text-driven motion generation.} Table \ref{tab:humanml results} and Table \ref{tab:kit results} present the results of text-driven motion generation on the HumanML3D and KIT-ML datasets, respectively. As shown in the first two rows of both tables, motion reconstruction with discrete pose codes achieves high fidelity and closely matches
ground-truth text-motion consistency, providing a strong foundation for motion generation. CoMo attains either the best or the second-best performance on HumanML3D across five metrics and ranks within the top three in six metrics on KIT. Compared to existing state-of-the-art methods, CoMo not only achieves competitive motion fidelity (FID) but also enhances motion diversity and consistency between generated motion and text descriptions (R-Precision, MM-DIST, Diversity). MModality measures the diversity of motions generated from the same text description. Due to the use of semantically meaningful pose codes, CoMo builds a stronger binding between text and the generated motion sequence, which potentially causes a slightly lower MModality score compared to prior methods as a trade-off for consistency. More importantly, our approach provides an intuitive interface for LLM-based zero-shot motion editing, which will be evaluated in Section \ref{sec: human evaluation} via a user study.

\begin{figure*}[!t]
    \centering
    \includegraphics[width=0.95\textwidth]{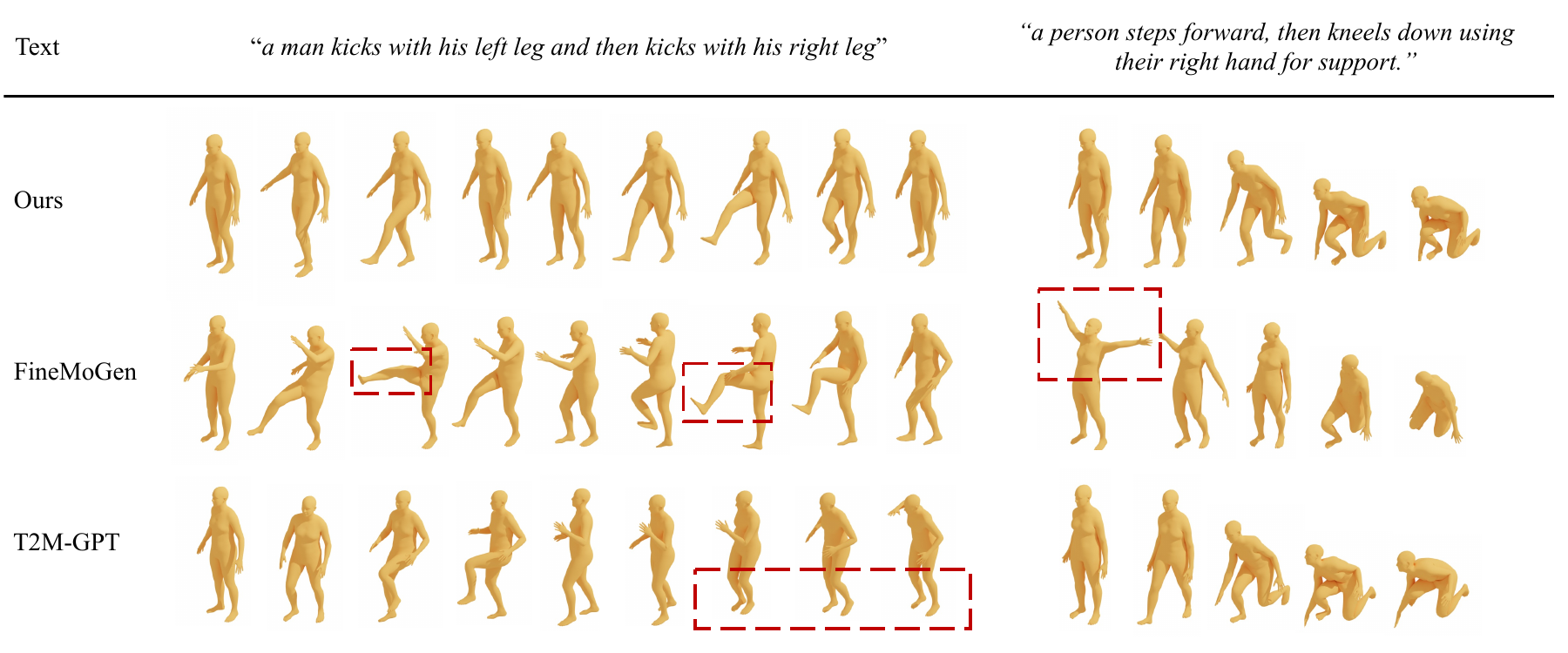}
    \caption{\textbf{Qualitative examples of Motion Generation on the HumanML3D test set \cite{t2m_guo}.} The motion sequences progress from left to right. The \textcolor{red}{\textbf{red}} boxes identify misalignments between the generated motion sequence and the text description. CoMo achieves competitive results in motion generation compared to T2M-GPT \cite{t2mgpt} and FineMoGen \cite{zhang2023finemogen}. More visual results are available in the Appendix.}
    \label{fig:gen_qualitative}
\end{figure*}

\subsection{Human Evaluation on Motion Editing} \label{sec: human evaluation}
We define the task of motion editing as the modification of a source motion according to a given textual editing instruction. Prior methods rely on modifying textual descriptions of the source motion based on edit instructions and subsequently generating new sequences from the updated descriptions to achieve motion editing \cite{zhang2023finemogen}. In contrast, CoMo distinguishes itself by directly interpreting and manipulating the source motion sequence to facilitate editing. 

\begin{figure}[!t]
    \centering
    \includegraphics[width=0.95\textwidth]{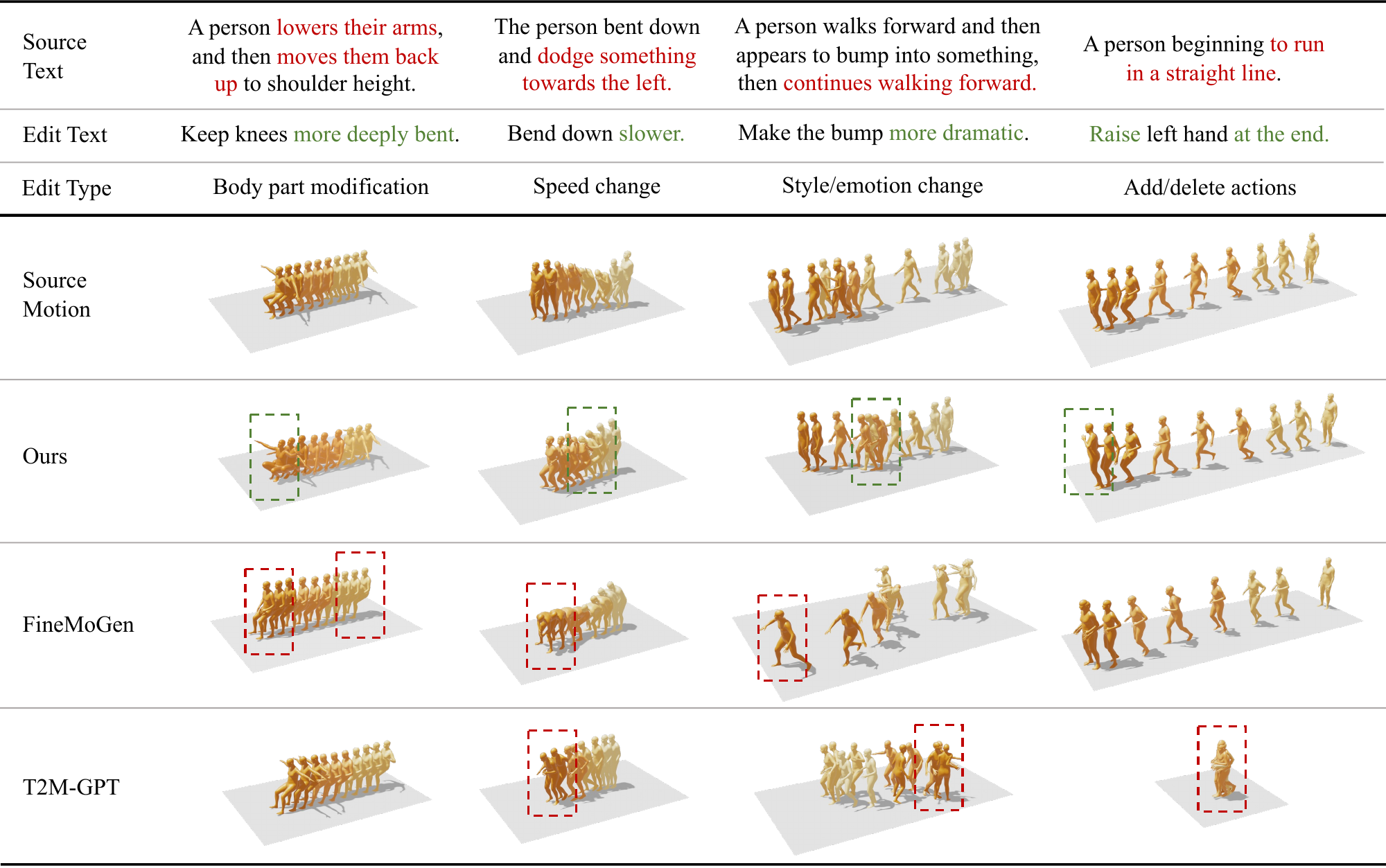}
    \caption{\textbf{Qualitative examples of Motion Editing on the HumanML3D test set \cite{t2m_guo}.} The {\color[HTML]{548235}\textbf{green}} words/boxes highlight successful edits. The \textcolor{red}{\textbf{red}} words/boxes identify misalignments between edited and source motions. Compared to other methods, CoMo achieves accurate edits while preserving key characteristics of the source motion.} 
    \label{fig:qualitative}
\end{figure}

We conducted a user study to assess the motion editing quality of CoMo in comparison with two state-of-the-art models for fine-grained text-to-motion generation: T2M-GPT \cite{t2mgpt} and FineMoGen \cite{zhang2023finemogen}. We randomly selected 20 examples from the HumanML3D test set, annotating each with a motion edit instruction. As illustrated in Figure \ref{fig:qualitative}, these edit instructions encompass four types of motion editing: 1) body part modification (e.g., \textit{"keep knees more deeply bent"}), 2) speed change (e.g., \textit{"bend down slower"}), 3) style/emotion change (e.g., \textit{"more dramatic"}) and 4) action addition/deletion (e.g., \textit{"raise left hand at the end"}).\\

\noindent \textbf{Baselines.} To form strong baselines, we prompt GPT-4 to generate an updated description in the style of HumanML3D annotations using the original description and edit instructions as context\footnote{Prompts and updated descriptions are available in the Appendix.}. Wcription to generate the edited motion for T2M-GPT and FineMoGen. For CoMo, we decompose the source motion as pose codes and prompt GPT-4 to edit the code sequence based on the original description, edit instruction and pose code semantics without fine-tuning. The edited code sequence is decoded to reconstruct the edited motion.\\

\begin{figure}[!t]
    \centering
    \includegraphics[width=0.8\textwidth]{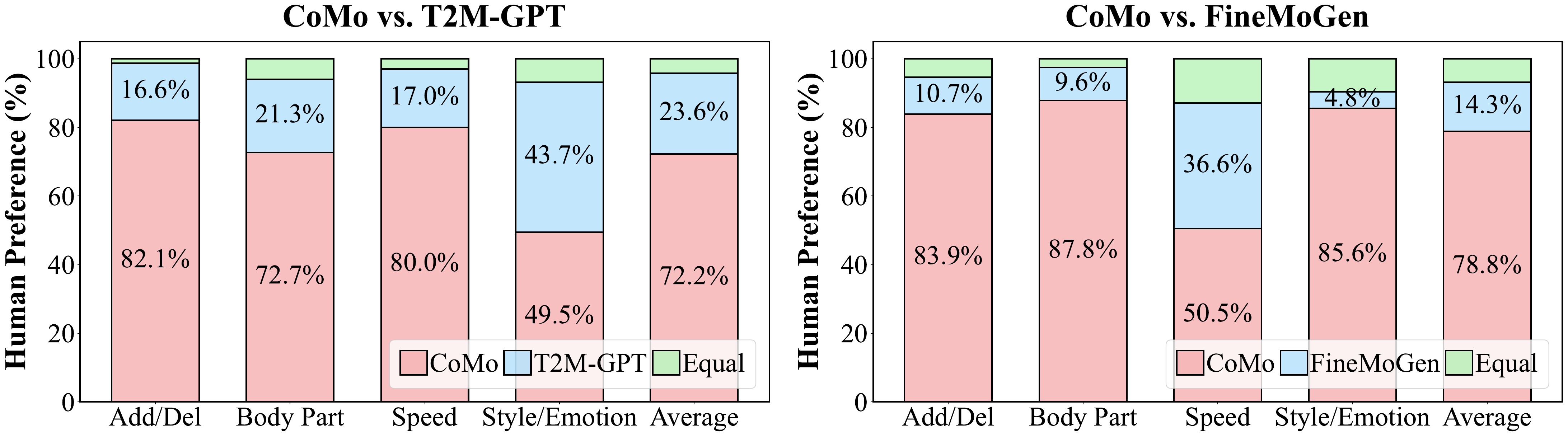}
    \caption{Human preference on Motion Editing by comparing CoMo with T2M-GPT \cite{t2mgpt} and FineMoGen \cite{zhang2023finemogen}. We report the scores on five editing types and average results.}
    \label{fig:human_eval}
\end{figure}

\noindent \textbf{User Study Setup.} We have 54 graduate students evaluate 20 editing scenarios, with 6 scenarios each for body part modification and add/delete actions edits and 4 scenarios each for speed change and style/emotion change. The edit categories are evenly divided between the two baselines. For each edit scenario, participants are shown a pair of edited motion sequences, one from a baseline model and one from CoMo, along with the source motion sequence, source motion description (e.g., ``a person is waving''), and an edit instruction (e.g., ``raise the left hand higher''). The users are tasked with comparing the edited motions and choosing which best reflects the provided instruction and preserves the characteristics of the source motion that are not affected by the edit.\\

\begin{table}[!t]
\centering
\caption{Ablation study of LLM-generated fine-grained keywords on HumanML3D and KIT. $-$Fine stands for the model without augmented keywords.}
\scriptsize
\resizebox{\textwidth}{!}{%
\begin{tabular}{lcccc|cccc}
\toprule
\multirow{2}{*}{\textbf{Method}} & \multicolumn{4}{c|}{\textbf{HumanML3D}}         & \multicolumn{4}{c}{\textbf{KIT Motion-Language}}          \\  \cmidrule{2-9}
                        & \textbf{Top-1 $\uparrow$} & \textbf{FID $\downarrow$} & \textbf{MM-DIST $\downarrow$} & \textbf{Diversity $\uparrow$} & \textbf{Top-1 $\uparrow$} & \textbf{FID $\downarrow$} & \textbf{MM-DIST $\downarrow$} & \textbf{Diversity $\uparrow$} \\ \midrule
\cellcolor{gray!10}CoMo           &   \cellcolor{gray!10}\textbf{0.502}    &   \cellcolor{gray!10}\textbf{0.262}  &    \cellcolor{gray!10}\textbf{3.032}     &    \cellcolor{gray!10}\textbf{9.936}            &  \cellcolor{gray!10}\textbf{0.422}    &   \cellcolor{gray!10}\textbf{0.332}       &  \cellcolor{gray!10}\textbf{2.873}          &   \cellcolor{gray!10}10.95         \\
$-$Fine &  0.487     &  0.263   &   3.044      &   9.519            &   0.399        &   0.399        &    2.898        &    \textbf{11.26}        \\      \bottomrule
\end{tabular}
}
\label{tab:finegrained}
\end{table}

\begin{table}[!t]
\centering
\caption{Ablation study of different codebook sizes $N$. We report the reconstruction performance on the HumanML3D validation set. The number of angle/distance cutoffs stands for the granularity of parsing the joint position.}
\scriptsize
\begin{tabular}{ccccccc}
\toprule
\multicolumn{1}{c}{\begin{tabular}[c]{@{}c@{}}\textbf{number of}\\ \textbf{codes ($N$)}\end{tabular}} &
\multicolumn{1}{c}{\begin{tabular}[c]{@{}c@{}}\textbf{angle}\\ \textbf{cutoffs}\end{tabular}} &
\multicolumn{1}{c}{\begin{tabular}[c]{@{}c@{}}\textbf{distance}\\ \textbf{cutoffs}\end{tabular}} &
\textbf{Top-1 $\uparrow$} & \textbf{FID $\downarrow$} & \textbf{MM-DIST $\downarrow$} & \textbf{Diversity $\uparrow$} \\ \midrule
205 & 3 & 2 & 0.500 & 0.059 & 3.030 & 9.592 \\
261 & 6 & 4 & 0.505 & 0.049 & 3.023 & 9.620 \\
\cellcolor{gray!10}392 & \cellcolor{gray!10}18 & \cellcolor{gray!10}10 & \cellcolor{gray!10}\textbf{0.517} & \cellcolor{gray!10}\textbf{0.034} & \cellcolor{gray!10}\textbf{2.770} & \cellcolor{gray!10}\textbf{10.030} \\
661 & 18 & 20 & 0.507 & 0.047 & 3.009 & 9.553 \\
733 & 36 & 20 & 0.504 & 0.037 & 3.019 & 9.520 \\
\bottomrule
\end{tabular}
\label{tab:codebook_size}
\end{table}

\begin{table}[!t]
\centering
\caption{Ablation study of different sampling rates $l$. We report the reconstruction performance on the HumanML3D validation set.}
\label{tab:sampling_rate}
\scriptsize
\begin{tabular}{ccccccc}
\toprule
\textbf{Sampling Rate ($l$)}         & \textbf{Top-1 $\uparrow$} & \textbf{Top-2 $\uparrow$} & \textbf{Top-3 $\uparrow$}& \textbf{FID $\downarrow$} & \textbf{MM-DIST $\downarrow$} & \textbf{Diversity $\uparrow$}  \\ \midrule
2           &   \textbf{0.523} & \textbf{0.723} & \textbf{0.821} & \textbf{0.021} & \textbf{2.751} & 10.048 \\
\cellcolor{gray!10}4           &   \cellcolor{gray!10}0.517 & \cellcolor{gray!10}0.718 & \cellcolor{gray!10}0.815 & \cellcolor{gray!10}0.034 & \cellcolor{gray!10}2.770 & \cellcolor{gray!10}10.030 \\
8           &  0.513 & 0.712 & 0.809 & 0.065 & 2.824 & \textbf{10.084} \\
16          &  0.486 & 0.682 & 0.782 & 0.176 & 3.021 & 10.039 \\
\bottomrule 
\end{tabular}
\end{table}

\noindent \textbf{Humans prefer CoMo for Motion Editing.} We present the average percentage of users' preferences for CoMo versus T2M-GPT and FineMoGen in Figure \ref{fig:human_eval}. The results show a clear preference (over 70\% on average) for motion editing with CoMo, especially in scenarios thatdify fine-grained motion details (e.g. body part modification and add/delete action). As depicted in Figure \ref{fig:qualitative}, while T2M-GPT and FineMoGen can produce motion relevant to the updated description, they often struggle to generate well-aligned details from scratch without any spatial-temporal context of the source motion. In contrast, CoMo leverages the semantics of pose codes to interpret the source motion effectively and achieve detailed motion editing. The observed decrease in preference for holistic motion edits involving changes in emotion or speed suggests that textual descriptions may provide more guidance in creating comprehensive edits that impact multiple aspects of a motion sequence. With its capability for high-level text-guided generation and iterative fine-grained pose code editing, CoMo emerges as a competitive approach for text-driven, controllable motion generation.

\subsection{Ablation} \label{sec: ablation}
This section ablates the fine-grained keywords, codebook size, and downsampling rate. Additional ablation studies are available in the Appendix.\\

\noindent\textbf{Fine-grained Keywords.} Table \ref{tab:finegrained} shows an ablation study on the role of LLM-generated fine-grained keywords conducted on the HumanML3D dataset. The performance of the model decreases without those keywords. The specific details introduced in fine-grained keywords allow for a more consistent mapping between the original text description and generated motion.\\ 

\noindent \textbf{Codebook Size.} The number of pose codes indicates how fine-grained the heuristic thresholds of the skeleton parser are. More fine-grained thresholds capture more details at the expense of computational speed and learning complexity. We evaluate motion reconstruction quality for different codebook sizes by varying the angle and distance threshold settings, with finer thresholds corresponding to larger codebooks. Results in Table \ref{tab:codebook_size} indicate that our current codebook setting of $392$ pose codes best balances complexity and reconstruction quality.\\ 

\noindent \textbf{Downsampling Rate.} We also investigate the effect of temporal downsampling on reconstruction quality. Similar to codebook size, smaller downsampling rates reserve details but result in longer sequences that increase task complexity and computation time. Comparing the results in Table \ref{tab:sampling_rate}, we select a downsampling rate of $4$ to balance model performance and complexity.

\section{Conclusion}
In this paper, we propose CoMo, a controllable human motion synthesis system capable of generating and editing motion through language inputs. CoMo adopts a semantically meaningful pose code representation for encoding motion sequences across space and time. The interpretable pose codes within CoMo enable large language models to understand motion sequences and perform both kinematic and semantic motion editing effectively in a zero-shot manner. CoMo achieves state-of-the-art results in text-driven motion generation and human preferences confirm its superiority over alternative systems for motion editing.\\

\noindent \textbf{Limitations.} Although CoMo enhances controllability through keywords and a motion editing interface,  the current formulation of keywords and pose codes focuses more on local kinematic descriptions. Further expanding the types of keywords and pose codes to include global descriptors of speed, style, trajectory, and motion repetition may allow for more flexibility in text-driven motion generation and editing. In addition. the semantics of pose codes enable zero-shot motion editing capabilities with LLMs but do not strictly constrain the motion edits to create physically feasible motion sequences. In the future, we aim to incorporate physical priors to guide the reasoning of LLMs during motion editing on pose codes to further enhance performance.

\clearpage
\section*{Acknowledgment}
This research is supported in part by the Office of the Director of National Intelligence (ODNI), Intelligence Advanced Research Projects Activity (IARPA), via the HIATUS Program contract \#2022-22072200005. The views and conclusions contained herein are those of the authors and should not be interpreted as necessarily representing the official policies, either expressed or implied, of ODNI, IARPA, or the U.S. Government. The U.S. Government is authorized to reproduce and distribute reprints for governmental purposes notwithstanding any copyright annotation therein.

\bibliographystyle{splncs04}
\bibliography{main}
\clearpage

\appendix
\appendix
\renewcommand\thefigure{\Alph{section}\arabic{figure}}    
\renewcommand\thetable{\Alph{section}\arabic{table}}
\section{Implementation Details}

\subsection{Model Architectures}

\begin{figure}[!ht]
    \centering
    \includegraphics[width=\textwidth]{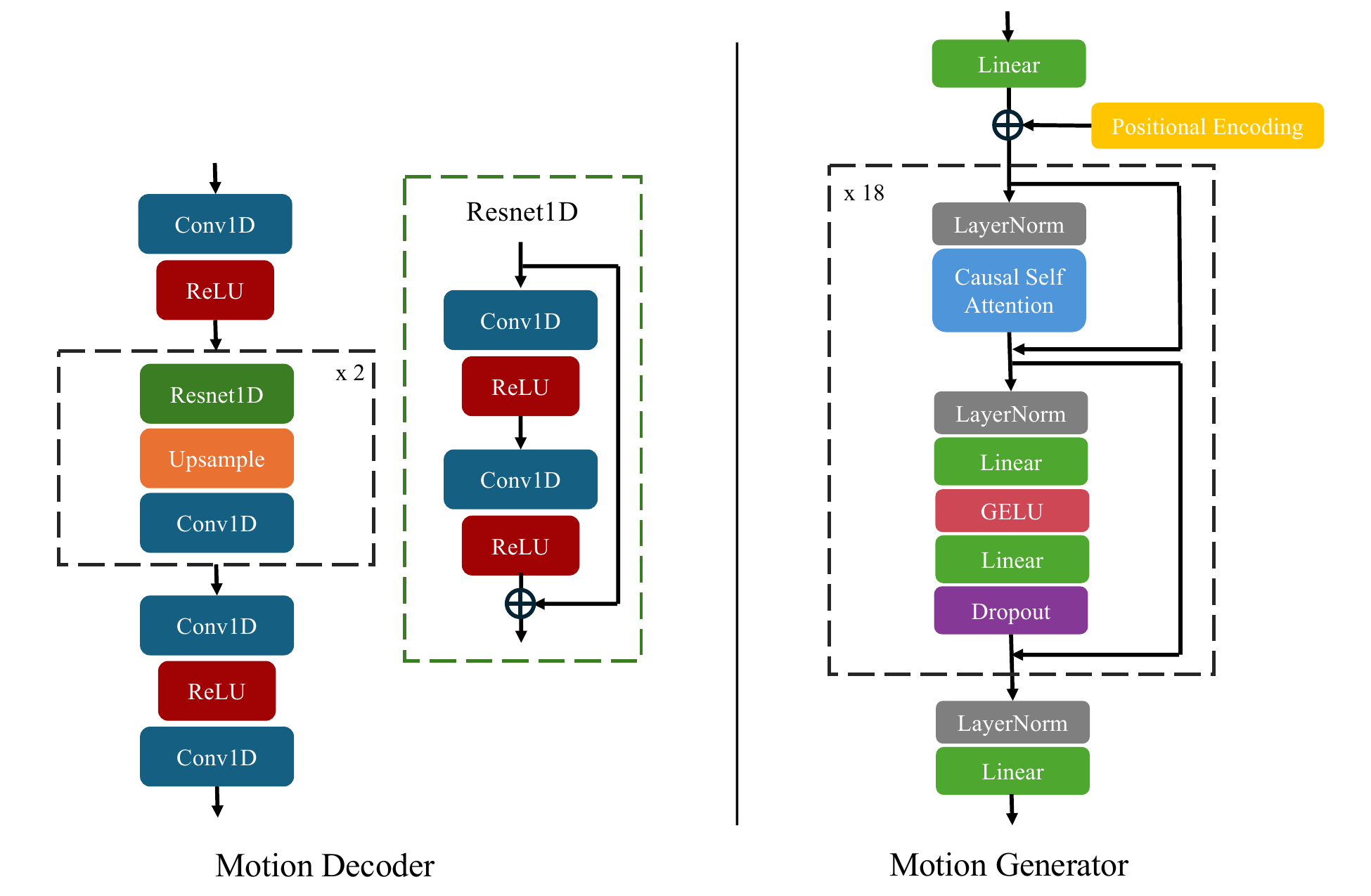}
    \caption{Architectures of the Motion Decoder and Motion Generator of CoMo.}
    \label{fig:model_architecture}
\end{figure}

We follow the setup in \cite{t2mgpt} for the decoder and transformer architectures. As Figure \ref{fig:model_architecture} illustrates, the Motion Decoder consists of 1D convolution layers, two residual blocks for upsampling, and ReLU activation layers. The codebook size is set to be $392 \times 512$. The Motion Generator uses a linear layer to project the K-hot pose code vectors before applying positional encoding and passing the sequence through a decoder-only transformer with causal self-attention blocks.
\subsection{Pose Codebook} We follow the overall computation method of pose codes designated by \cite{delmas2022posescript} and adjust the heuristics to accommodate the task of motion generation. We select 70 pose code categories consisting of 392 pose codes. For each pose code category, a corresponding type of heuristic threshold is applied to define the pose codes within each category (e.g., ``L-knee angle'' will have a set of angle thresholds). Table \ref{tab:code_thresh} shows the 7 different types of heuristic thresholds used. Amongst the 70 pose code categories, 4 use the angle threshold, 18 use the distance threshold, 31 use relative position thresholds (6 along the x-axis, 16 along the y-axis, 9 along the z-axis), 13 use the relative orientation threshold, and 4 use the ground-contact threshold. Table \ref{tab:code_cat} shows a list of the 70 pose code categories grouped according to the type of threshold they use. The semantics of pose codes are defined by combining the pose code category names, which indicate the joints involved, with the joint states described by the threshold conditions.
\medbreak
\noindent \textbf{Mutual Exclusivity} Although we frame our optimization objective concerning independent Bernoulli variables, certain pose codes can be mutually exclusive. For example, the concepts ``L-arm below torso'' and ``L-arm above torso'' are contradictory and should not co-occur. The encoded motion sequence is structured as $K$-hot, representing $K$ pose code categories, each corresponding to a subset of mutually exclusive codes. During training, mutual exclusivity is ensured within the ground truth representation. During inference, we enforce this mutual exclusivity for each of the $K$ categories by activating the pose code with the highest log-likelihood within its corresponding category. We can also generate diverse motion sequences by sampling from the predicted distributions of the $K$ categories provided by the transformer model (see Figure \ref{fig:diversity}).
\medbreak
\noindent \textbf{Sequence Corruption.} To mitigate the discrepancy between training and testing, we randomly replace the tokens within each subset during training while maintaining mutual exclusivity. Additionally, we perform random masking on the fine-grained, part-specific text conditions to facilitate effective text-based control.

\begin{table}[!ht]
\centering
\caption{\textbf{Pose Code Categories.} A total of 70 categories of pose codes are used, grouped according to their threshold type. We denote Left and Right as `L' and `R' respectively. The axes considered for relative positions are noted in parentheses.}
\resizebox{\textwidth}{!}{%
\begin{tabular}{ccccc}
\toprule
\textbf{Angle} & \textbf{Distance} & \textbf{Relative Position} & \textbf{Relative Orientation} & \textbf{Ground-contact} \\ \hline
L-knee & L-elbow vs R-elbow & L-shoulder vs R-shoulder (YZ) & L-hip vs L-knee & L-knee \\
R-knee & L-hand vs R-hand & L-elbow vs R-elbow (YZ) & R-hip vs R-knee & R-knee \\
L-elbow & L-knee vs R-knee & L-hand vs R-hand (XYZ) & L-knee vs L-ankle & L-foot \\
R-elbow & L-foot vs R-foot & neck vs pelvis (XZ) & R-knee vs R-ankle & R-foot \\
 & L-hand vs L-shoulder & L-ankle vs neck (Y) & L-shoulder vs L-elbow & \\
 & L-hand vs R-shoulder & R-ankle vs neck (Y) & R-shoulder vs R-elbow & \\
 & R-hand vs L-shoulder & L-hip vs L-knee (Y) & L-elbow vs L-wrist & \\
 & R-hand vs R-shoulder & R-hip vs R-knee (Y) & R-elbow vs R-wrist & \\
 & L-hand vs R-elbow & L-hand vs L-shoulder (XY) & pelvis vs L-shoulder & \\
 & R-hand vs L-elbow & R-hand vs R-shoulder (XY) & pelvis vs R-shoulder & \\
 & L-hand vs L-knee & L-foot vs L-hip (XY) & pelvis vs neck & \\
 & L-hand vs R-knee & R-foot vs R-hip (XY) & L-hand vs R-hand & \\
 & R-hand vs L-knee & L-wrist vs neck (Y) & L-foot vs R-foot & \\
 & R-hand vs R-knee & R-wrist vs neck (Y) &  & \\
 & L-hand vs L-foot & L-hand vs L-hip (Y) &  & \\
 & L-hand vs R-foot & R-hand vs R-hip (Y) &  & \\
 & R-hand vs L-foot & L-hand vs torso (Z) &  & \\
 & R-hand vs R-foot & R-hand vs torso (Z) &  & \\
 & &L-foot vs torso (Z) & &\\
 & &R-foot vs torso (Z) & &\\
 & &L-knee vs R-knee (YZ) & &\\
\bottomrule 
\end{tabular}
}
\label{tab:code_cat}
\end{table}

\begin{table}[!htbp]
\small
\centering
\begin{tabular}{ccc}
\toprule
\textbf{Threshold Type}   & \textbf{Semantics} & \textbf{Threshold Condition}  \\ \hline
\multirow{18}{*}{Angle} & bent to almost 10 degrees & $  x \leq 10$ \\
& bent to almost 20 degrees & $10 <x\leq  <20$ \\
& bent to almost 30 degrees & $20 <x\leq 30$ \\
& bent to almost 40 degrees & $30 <x\leq 40$ \\
& bent to almost 50 degrees & $40 <x\leq 50$ \\
& bent to almost 60 degrees & $50 <x\leq 60$ \\
& bent to almost 70 degrees & $60 <x\leq 70$ \\
& bent to almost 80 degrees & $70 <x\leq 80$ \\
& bent to almost 90 degrees & $80 <X\leq 90$ \\
& bent to almost 100 degrees & $90 <x\leq 100$ \\
& bent to almost 110 degrees & $100 <x\leq 110$ \\
& bent to almost 120 degrees & $110 <x\leq 120$ \\
& bent to almost 130 degrees & $120 <x\leq 130$ \\
& bent to almost 140 degrees & $130 <x\leq 140$ \\
& bent to almost 150 degrees & $140 <x\leq 150$ \\
& bent to almost 160 degrees & $150 <x\leq160$ \\
& bent to almost 170 degrees & $160 <x\leq 170$ \\
& straight &  $x >170$ \\
\hline 
\multirow{10}{*}{Distance} & very close& $  x \leq 0.1$ \\
& slightly close & $0.1 <X\leq 0.2$ \\
& close & $0.2 <x\leq  0.3$ \\
& almost shoulder width apart & $0.3 <x\leq 0.4$ \\
& shoulder with apart & $0.4 <x\leq 0.5$ \\
& almost spread & $0.5 <x\leq 0.6$ \\
& spread & $0.6 <x\leq 0.7$ \\
& slightly wide & $0.7 <x\leq 0.8$ \\
& wide & $0.8 <x\leq <0.9$ \\
& very wide & $ x > 0.9$\\
\hline
\multirow{3}{*}{Relative Position along X axis} & at the right of & $  x \leq -0.15$ \\
& ignored & $  -0.15 < x \leq 0.15$ \\
& at the left of & $  x > 0.15$ \\
\hline
\multirow{3}{*}{Relative Position along Y axis} & below & $  x \leq -0.15$ \\
& ignored & $  -0.15 < x \leq 0.15$ \\
& above & $  x > 0.15$ \\
\hline
\multirow{3}{*}{Relative Position along Z axis} & behind & $  x \leq -0.15$ \\
& ignored & $  -0.15 < x \leq 0.15$ \\
& in front of & $  x > 0.15$ \\
\hline
\multirow{3}{*}{Relative Orientation} & vertical & $  x \leq 10$ \\
& ignored & $  10 < x \leq 80$ \\
& horizontal & $  x \leq 80$ \\
\hline
\multirow{2}{*}{Ground-contact} & on the ground & $  x \leq 0.1$ \\
& ground-ignored & $  x > 0.1$ \\
\bottomrule 
\end{tabular}
\vspace{0.1cm}
\caption{\textbf{Pose Code Threshold Conditions.} For each pose code category, a corresponding threshold type is applied to specify the semantics of pose codes within that category. $x$ represents the input value. Angles are represented in degrees, distances/relative positions/ground contact are represented in meters, and relative orientation is represented by the cosine similarity between unit vectors along the y-axis.}
\vspace{-0.3cm}
\label{tab:code_thresh}
\end{table}

\subsection{Fine-grained Keyword Setup}
Figure \ref{fig:prompt-keywords} presents the prompt template used to interact with GPT-4 to generate fine-grained keywords that enhance the text descriptions for the motion sequences. For each motion sample, we generate 5 different sets of fine-grained keywords using the original text description as input. During training, the keywords paired with the motion samples are chosen randomly from these five sets. An example of a text description from the HumanML3D training set \cite{t2m_guo} and the corresponding generated fine-grained keywords is presented in Table \ref{tab:keywords}. The generated keywords expand the original text description to include additional details that depict the motion of specific body parts. More examples of text descriptions, generated fine-grained keywords, and the corresponding generated motion sequence are shown in Figure \ref{fig:diversity}.
\begin{figure}[h]
    \centering
    \begin{center}
    \begin{tcolorbox} [top=3pt,bottom=3pt, left=3pt, right=3pt, width=\linewidth, boxrule=1pt]
    {\scriptsize {\fontfamily{zi4}\selectfont    
    Given a text description of a motion: \textbf{\{details\}}. Enrich the description of the full motion by summarizing in detail the shape and speed for each of the body parts in \textbf{\{body\_parts\}} that is required to achieve the given motion in natural language. The output should be in json format with \textbf{\{body\_parts\}} as keys, and one short motion attribute as values. Key-value format example: "head":"head is upright". Do not output anything else.}    \par}
    \end{tcolorbox}
    \begin{tcolorbox} [top=3pt,bottom=3pt, left=3pt, right=3pt, width=\linewidth, boxrule=1pt]
    {\scriptsize {\fontfamily{zi4}\selectfont    
    Given a text description of a motion: 
    Please help me to describe the mood that is required to achieve the human motion described as: '\textbf{\{details\}}' using one short motion attribute. Do not output anything else.}    \par}
    \end{tcolorbox}
    \end{center}
    \vspace{-0.2cm}
    \caption{Prompt template for generating fine-grained keywords.}
    \label{fig:prompt-keywords}
\end{figure}
\begin{table}[!ht]
\renewcommand{\arraystretch}{1.8} 
\centering
\caption{An example of generated fine-grained keywords on HumanML3D.}
\scriptsize
\resizebox{\textwidth}{!}{
\begin{tabular}
{|p{1.8cm}|p{1.5cm}|p{9cm}|}
\hline
\multicolumn{2}{|c|}{\textbf{Text}} & \multicolumn{1}{l|}{a man stumbles to his right} \\ \hline
\multirow{11}{*}{\begin{tabular}[c]{@{}l@{}} Fine-grained\\keywords \end{tabular}} & Head & The man's head tilts slightly forward and turns to the right, following the direction of the stumble \\ \cline{2-3} 
 & Torso & The man's torso leans to the right, as if losing balance. \\ \cline{2-3} 
 & L-Arm & The man's left arm swings outward to the left, in an instinctive attempt to regain balance \\ \cline{2-3} 
 & R-Arm & The man's right arm tucks in towards his body as he stumbles to the right \\ \cline{2-3} 
 & L-Hand & The man's left hand is open, ready to grasp anything in the vicinity for support if needed \\ \cline{2-3} 
 & R-Hand & The man's right hand clenches slightly, moving in unison with the right arm\\ \cline{2-3} 
 & L-Leg & The man's left leg is firmly planted, acting as the pivot for the stumble\\ \cline{2-3} 
 & R-Leg  & The man's right leg lifts and steps awkwardly to the right, causing the stumble\\ \cline{2-3} 
 & L-Feet & The man's left foot remains grounded, providing the only source of stability during the stumble\\ \cline{2-3} 
 & R-Feet & The man's right foot lands unevenly on the ground, leading to the stumble\\ \cline{2-3} 
 & Mood & Unsteady\\ \cline{2-3} \hline
\end{tabular}}
\label{tab:keywords}
\end{table}
\subsection{Metrics.} For quantitative evaluation, we obtain motion and text feature vectors using the motion feature extractor and text feature extractor pretrained in \cite{t2m_guo} and calculate the following metrics:
\medbreak
\noindent \textit{FID} evaluates the quality of generated motion by computing the difference in the mean and covariance of the motion features.
\medbreak
\noindent \textit{R-Precision} measures the consistency between text and generated motion. The ground truth text description and a set of mismatched text descriptions are selected for each generated motion to form a pool. The descriptions in the pool are ranked based on the Euclidean distance between their text feature and the generated motion feature, with smaller distances ranking higher. R-precision (Top-k) calculates the average probability of the ground truth text description ranking within the Top-k candidates.
\medbreak
\noindent \textit{MM-Dist} computes the average Euclidean distance between a text feature and the corresponding generated motion feature over $N$ randomly generated pairs.
\medbreak
\noindent \textit{Diversity} computes the average Euclidean distance between pairs of generated motion features over $M$ randomly generated pairs and indicates the variance of generated motion.
\medbreak
\noindent \textit{MModality} determines the diversity of generated motion for the same text condition. For each text description, several motions are generated, and the average distance between the generated motion features is computed. This value is then averaged across all text descriptions.
\medbreak
\subsection{Motion Editing Prompts} Figure \ref{fig:prompt-frame} and Figure \ref{fig:prompt-edit} present the prompt templates used to interact with GPT-4 for motion editing on pose codes. Pose code semantics are provided as a table to provide context for the LLM to interpret encoded motion sequences. 
\begin{figure}[!ht]
    \centering
    \begin{center}
    \begin{tcolorbox} [top=3pt,bottom=3pt, left=3pt, right=3pt, width=\linewidth, boxrule=1pt]
    {\scriptsize {\fontfamily{zi4}\selectfont
    
    Motion is represented by a set of joint states, defined as follows: \\
Table 1 Joint State Meanings (Key: Joint State Index, Value: Joint State Meaning):  \textbf{\{table1\}} \\
Given the edit instruction: \textbf{\{edit\}} \\
Return a semi-colon separated sequence of the ids of the joint states you will need to examine in order to determine the starting and ending frame of a motion sequence that will be affected by the edit instruction. \\ Format example: 0;1;5;9. Do not reply anything else.
}
    \par}
    \end{tcolorbox}
    \begin{tcolorbox} [top=3pt,bottom=3pt, left=3pt, right=3pt, width=\linewidth, boxrule=1pt]
    {\scriptsize {\fontfamily{zi4}\selectfont
    
You will be provided with a text description of the motion, a motion code sequence and a motion edit instruction. You are be required to determine the starting and ending frame of the sequence that will be affected by the edit. Here is what you need to know about the encoding of the motion sequences:\\
    The motion is represented a number of time frames, each time frame contains a set of joint states, each joint state contains a code value. The definitions are:\\
    Table 1 Joint State Meanings (Key: Joint State Index, Value: Joint State Meaning):  \textbf{\{table1\}}\\
    Table 2 Code Meaning (Key: Code ID, Value: Code Meaning): \textbf{\{table2\}}\\
    Rules: smaller angles indicates more bending.\\
    The motion code sequence is: \textbf{\{codes\}}\\
    The total number of time frames is \textbf{\{length\}}\\
    The text description is: \textbf{\{details\}}\\
    The edit instruction is: \textbf{\{edit\}}\\
    Return the starting index and ending index of the segment that is affected by the edit, separated by semi-colon, if the edit affects the overall movement, select the entire sequence. Format example: 0;19. Do not reply anything else.
}

    \par}
    \end{tcolorbox}
    
    \end{center}
    \vspace{-0.2cm}
    \caption{Prompt template for identifying the frames for editing.}
    \label{fig:prompt-frame}
\end{figure}
\begin{figure}[!t]
    \centering
    \begin{center}
    \begin{tcolorbox} [top=3pt,bottom=3pt, left=3pt, right=3pt, width=\linewidth, boxrule=1pt]
    {\scriptsize {\fontfamily{zi4}\selectfont
    
    Motion is represented by a set of joint states, defined as follows:\\
Table 1 Joint State Meanings (Key: Joint State Index, Value: Joint State Meaning):  \textbf{\{table1\}} \\
Given the edit instruction: \textbf{\{edit\}}\\
Return a semi-colon separated sequence of the ids of the joint states you may be affected by the edit instruction. Format example: 0;1;5;9. Do not reply anything else.

}
    \par}
    \end{tcolorbox}
    \begin{tcolorbox} [top=3pt,bottom=3pt, left=3pt, right=3pt, width=\linewidth, boxrule=1pt]
    {\scriptsize {\fontfamily{zi4}\selectfont
    
You will be provided with a text description of the motion, a motion code sequence for a given joint state and a motion edit instruction. You will be required to determine how to modify the codes within the provided sequence accordingly. \\
Here is what you need to know about the encoding of the motion sequences:
The motion is represented as a list of joint states of length T, T is the number time frames. Each joint state contains a code value. The usable codes are defined as follows: \\
Table 1 Usable Code Meaning (Key: Code ID, Value: Code Meaning): \textbf{\{table2\}} \\
Rules: smaller angles indicates more bending. \\
You are given this motion code sequence for the joint state \textbf{\{joint\}}, it has already been sliced to keep only the segment you will need to edit: \textbf{\{codes\}}. \\
The text description of the overall motion sequence is: \textbf{\{details\}}. \\
The edit instruction is: \textbf{\{edit\}} \\
Return the edited motion only as a sequence of integer code ids of length \textbf{\{length\}} separated by semi-colons, only use code ids in the provided table. If no edit needs to be made, return the original sequence. Format example: 1;2;3;4. Do not reply anything else. No explanation needed.

}

    \par}
    \end{tcolorbox}

    \end{center}
    \vspace{-0.2cm}
    \caption{Prompt template for identifying the body parts/joints for editing (above) and the prompt for executing the edits (below).}
    \label{fig:prompt-edit}
\end{figure}
\subsection{Inference Time}
\noindent We follow the metric in \cite{MLD} and calculate the Average Inference Time per Sentence (AITS) of our approach and T2M-GPT \cite{t2mgpt}, which has a similar architecture setup on the test set of HumanML3D \cite{t2m_guo}. AITS corresponds to the time cost in seconds for a model to generate one motion sequence, excluding the time costs for model and dataset loading. For our model, the time cost would cover both the generation of the pose code sequence and the decoding of pose codes into the final motion sequence. As shown in Table \ref{tab:time}, despite requiring longer sequence generation due to the addition of fine-grained keywords, our approach achieves new motion editing capabilities while maintaining competitive motion generation performance with marginal inference time increment.
\begin{table}[!htbp]
\small
\centering
\caption{Comparison of the Average Inference Time per Sentence costs on HumanML3D Test Set \cite{t2m_guo} with NVIDIA RTXA6000 GPU.}
\begin{tabular}{cc}
\toprule
\textbf{Methods}         & \textbf{AITS (s) $\downarrow$} \\ \midrule
Ours & 0.62\\
T2M-GPT \cite{t2mgpt} & 0.35\\
\bottomrule 
\end{tabular}
\label{tab:time}
\end{table}


\section{Human Evaluation}
\begin{figure}[!t]
    \centering
    \includegraphics[width=7cm]{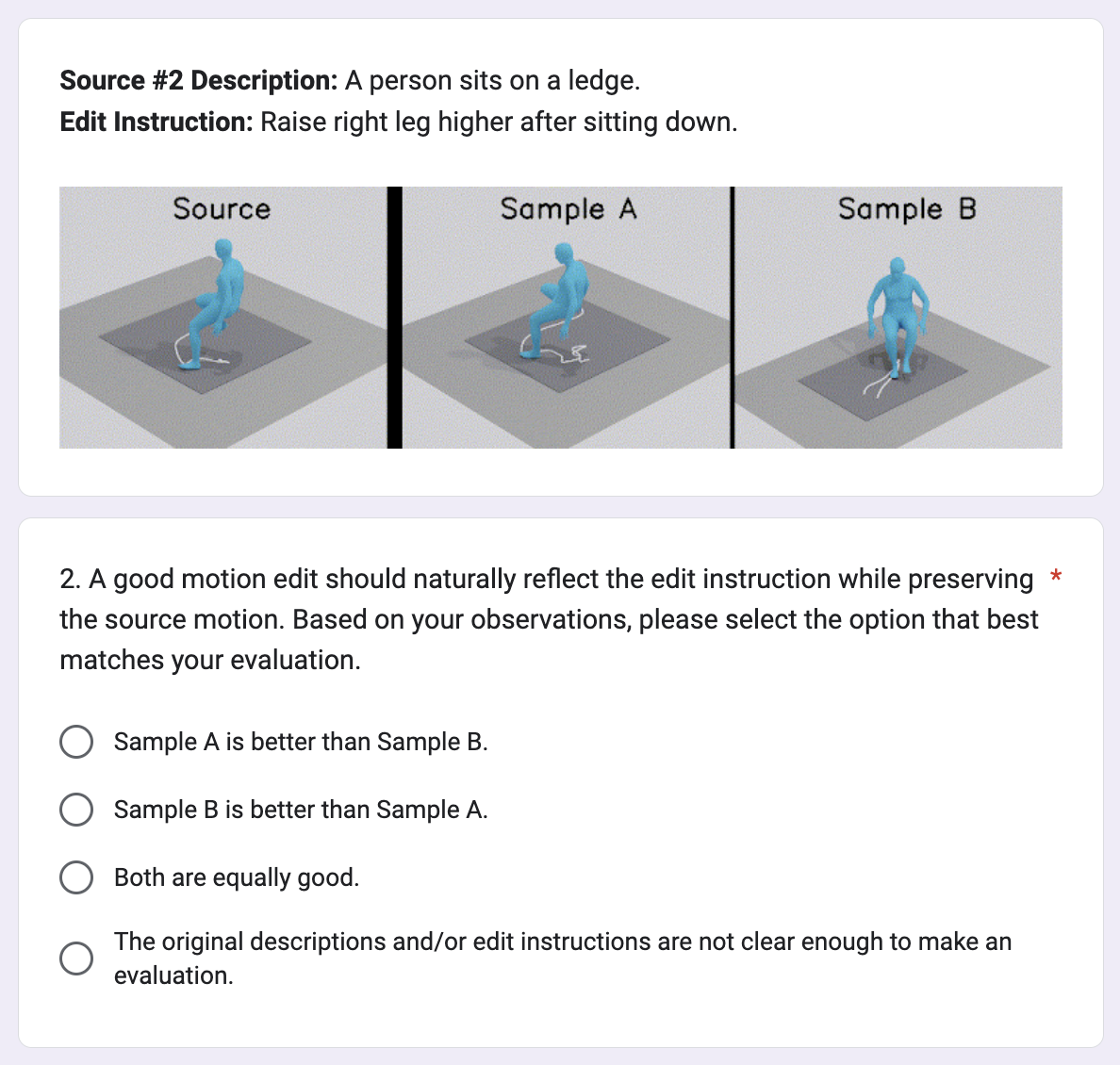}
    \caption{An example of our annotation interface in our user study. The option ``Both are equally good'' indicates the case with no clear advantage over the baselines, viewed as a negative evaluation during analysis.}
    \label{fig:annotation_interface}
\end{figure}

Figure \ref{fig:annotation_interface} shows a screenshot of our annotation interface. Users are provided with the source description and motion, the edit instruction, and the two edited motions from two different systems. We use the following prompt template to obtain updated descriptions for the baselines to generate edited motions:
\begin{tcolorbox} [top=3pt,bottom=3pt, left=3pt, right=3pt, width=\linewidth, boxrule=1pt]
{\scriptsize {\fontfamily{zi4}\selectfont
Given a source motion description: \textbf{\{details\}} and an edit instruction: \textbf{\{edit\}}, provide an updated motion description that describes the motion after applying the edit. The updated description should be similar in style as \textbf{\{examples\}}. Do not reply anything else.
}
\par}
\end{tcolorbox}
\noindent 10 randomly chosen text descriptions from the training set are provided as examples to guide the LLM to generate updated text descriptions in a similar style to the dataset annotations. Examples of the updated text descriptions used by the baselines are shown in Table \ref{tab:update_text}.
\begin{table}[!htbp]
\scriptsize
\renewcommand{\arraystretch}{1.5} 
\centering
\caption{Example updated descriptions for baseline motion editing.}
\resizebox{\textwidth}{!}{
\begin{tabular}{|p{4cm}|p{3cm}|p{5cm}|}
\hline
\textbf{Text Description} & \textbf{Edit Instruction} & \textbf{Updated Description} \\
\hline
A person lowers their arms, and then moves them back up to shoulder height & Keep both knees deeply bent & A person, with both knees deeply bent, lowers their arms and then raises them back up to shoulder height \\
\hline
The person bent down and dodge something towards the left & Bend down slower & The person slowly bent down and dodged something towards the left\\
\hline
A person walks forward and then appears to bump into something, then continues walking forward & Make the bump more dramatic & A person walks forward, then suddenly collides with a large unseen obstacle with a significant impact, recoiling notably before resuming their forward motion \\
\hline
A person beginning to run in a straight line & Raise left hand at the end & A person begins to run in a straight line and raises their left hand near the end of the run \\
\hline
\end{tabular}}
\label{tab:update_text}
\end{table}

\noindent A total of 54 graduate students participated in this user study. The Fleiss' kappa measurement for annotation agreement is 0.4, indicating moderate agreement among the raters.

\section{Additional Qualitative Examples}

\subsection{Motion Generation Examples}
\begin{figure}[!htpb]
    \centering
    \includegraphics[width=0.98\textwidth]{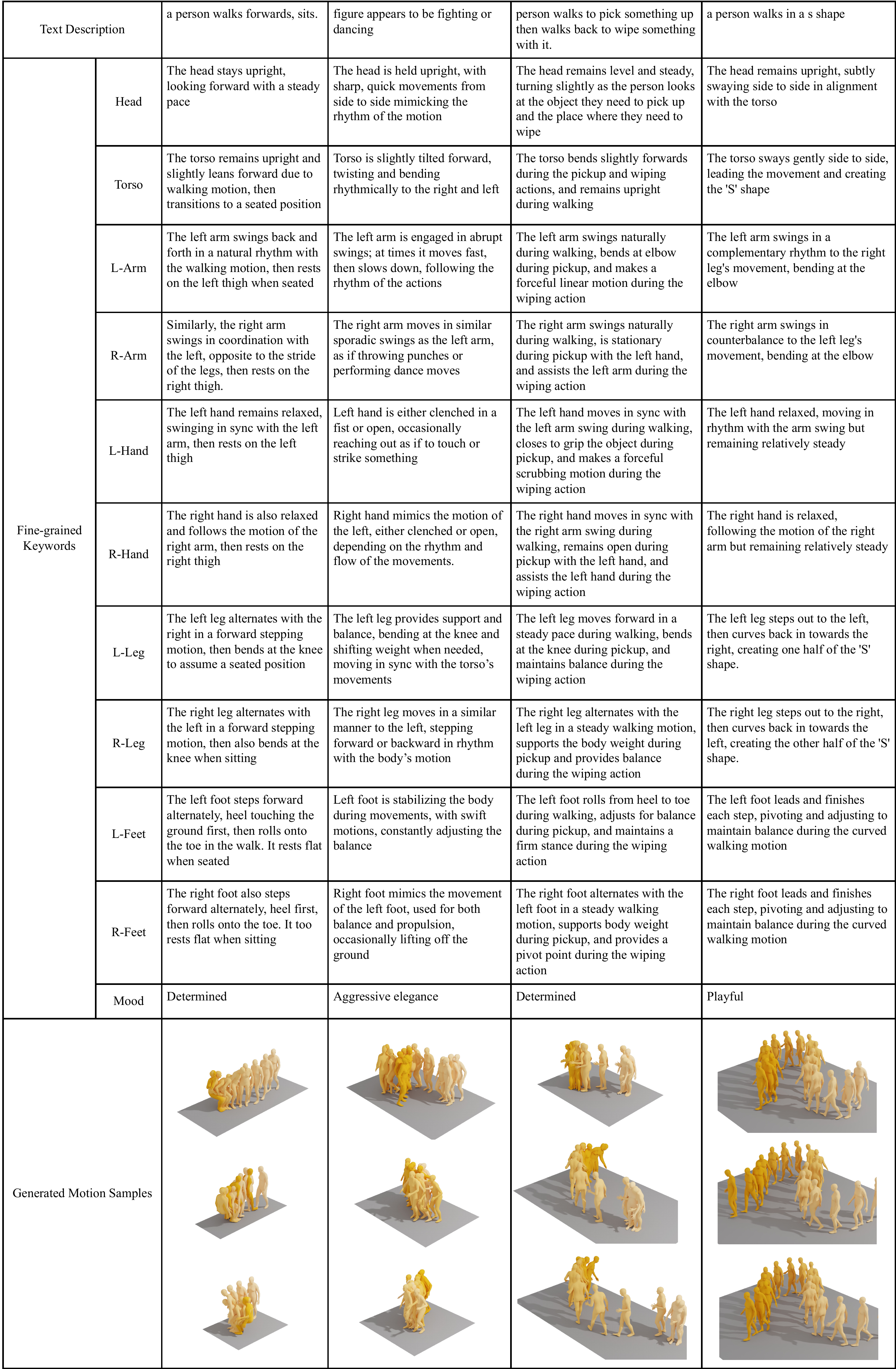}
    \caption{Qualitative examples of diverse motion generation.}
    \label{fig:diversity}
\end{figure}

In \ref{fig:diversity}, we generate three motion samples for each text description under the same text condition with fine-grained keywords from the HumanML3D test set. The results demonstrate the diversity of the generated motion and consistency between the motion and text conditions.

\subsection{Motion Editing Examples}

\begin{figure}[!htpb]
    \centering
    \includegraphics[width=\textwidth]{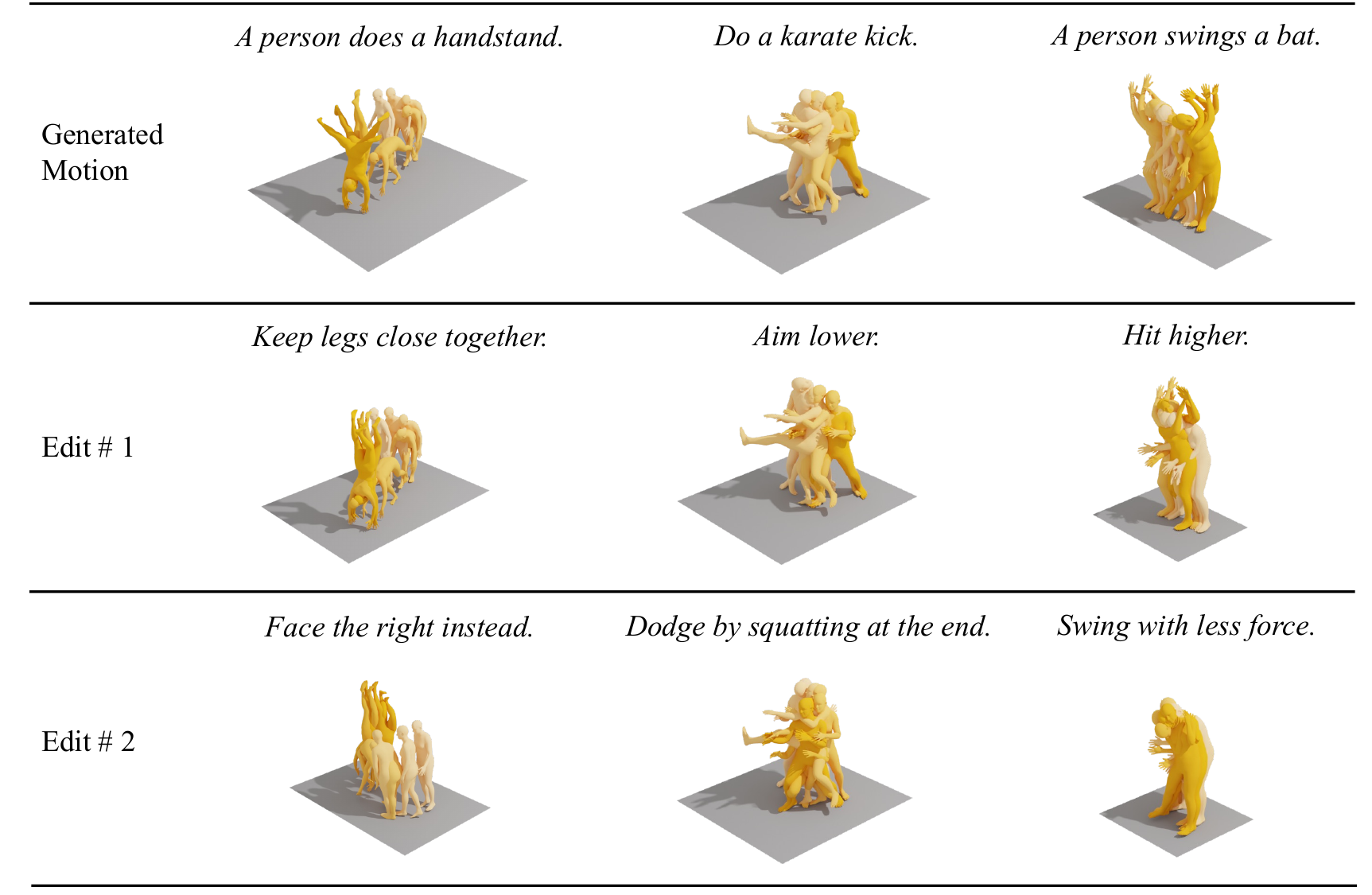}
    \caption{Qualitative examples of iterative motion editing. The Motion Generator generates the initial motion sequence using the provided text description. The Motion Editor then iteratively edits the pose code sequence based on edit instructions with previous edits preserved.}
    \label{fig:iterative}
\end{figure}

In \ref{fig:iterative}, we present qualitative examples for iterative motion editing, where a motion sequence is first generated from a text condition and then is edited iteratively by two edit instructions. The results demonstrate the capability of our approach for continuously interpreting and editing motion sequences, enabling both effective motion edits and the preservation of useful motion characteristics from previous iterations.

\begin{figure}[!htpb]
    \centering
    \includegraphics[width=\textwidth]{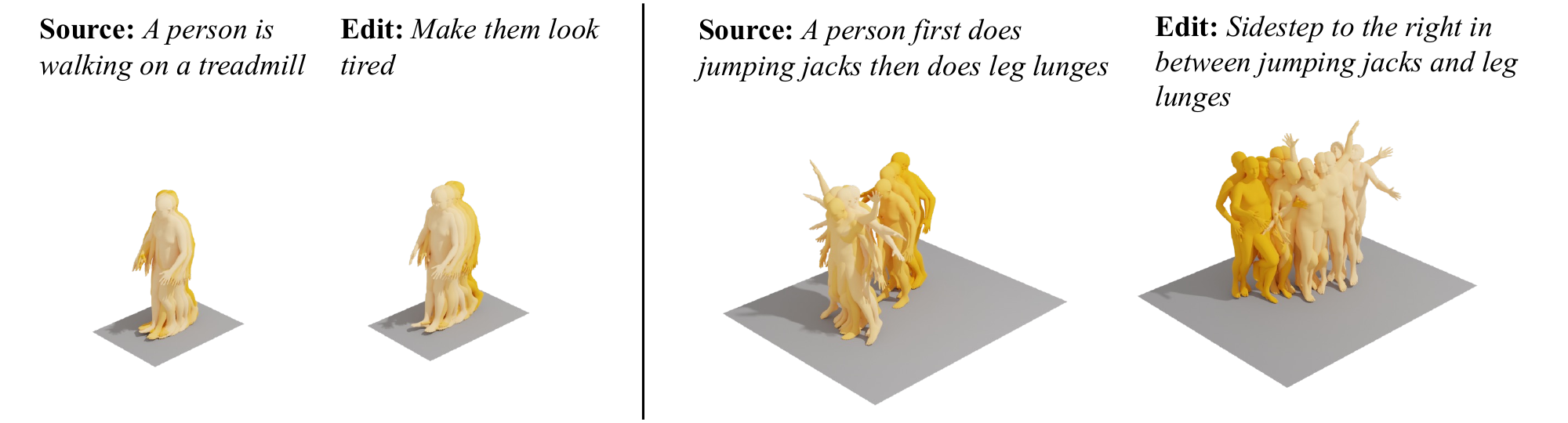}
    \caption{Failure cases in motion editing. \textbf{Left:} The edited motion does not depict the target emotion adequately. \textbf{Right:} The edited motion mistakenly added the 'sidestep' near the start of the motion rather than in between the two exercises.}
    \label{fig:fail}
\end{figure}

\medbreak
\noindent \textbf{Failure Cases:} As shown in \ref{fig:fail}, the semantics of pose codes focus on local kinematic attributes, which provides helpful context for LLMs to edit local joint states. However, for edits that require global changes in emotion or speed, the LLM may struggle with interpreting how the global edit translates to fine-grained modifications of local attributes. In addition, for more complex motion sequences with faster movement, the LLM tends to choose a broader range of frames when determining which frames to edit, which may limit the precision of the edit being made. In such cases, the user may intervene and directly select the time frames they want to edit.

\end{document}